\documentclass{article}

\usepackage{microtype}
\usepackage{graphicx}
\usepackage{subfigure}
\usepackage{booktabs} 

\usepackage{hyperref}

\usepackage{subfigure}
\usepackage{makecell}
\usepackage{multirow}
\usepackage{pifont}

\usepackage{bbm}
\usepackage{mathtools}

\usepackage[accepted]{icml2025}

\usepackage{amsmath}
\usepackage{amssymb}
\usepackage{mathtools}
\usepackage{amsthm}

\usepackage[capitalize,noabbrev]{cleveref}

\theoremstyle{plain}

\theoremstyle{definition}

\theoremstyle{remark}

\usepackage[textsize=tiny]{todonotes}

\newcommand{\ourmethod}{DriveGPT}

\icmltitlerunning{\ourmethod: Scaling Autoregressive Behavior Models for Driving}

\begin{document}

\twocolumn[
\icmltitle{\ourmethod: Scaling Autoregressive Behavior Models for Driving}



\icmlsetsymbol{equal}{*}

\begin{icmlauthorlist}
\icmlauthor{Xin Huang}{Cruise}
\icmlauthor{Eric~M.~Wolff}{Cruise}
\icmlauthor{Paul Vernaza}{Cruise}
\icmlauthor{Tung Phan-Minh}{Cruise}
\icmlauthor{Hongge Chen}{Cruise}
\icmlauthor{David~S.~Hayden}{Cruise}
\icmlauthor{Mark Edmonds}{Cruise}
\icmlauthor{Brian Pierce}{Cruise}
\icmlauthor{Xinxin Chen}{Cruise}
\icmlauthor{Pratik Elias Jacob}{Cruise}
\icmlauthor{Xiaobai Chen}{Cruise}
\icmlauthor{Chingiz Tairbekov}{Cruise}
\icmlauthor{Pratik Agarwal}{Cruise}
\icmlauthor{Tianshi Gao}{Cruise}
\icmlauthor{Yuning Chai}{Meta}
\icmlauthor{Siddhartha~S.~Srinivasa}{Cruise}
\end{icmlauthorlist}

\icmlaffiliation{Cruise}{Cruise LLC, San Francisco, CA}
\icmlaffiliation{Meta}{Meta, Menlo Park, CA}

\icmlcorrespondingauthor{Xin Huang, Eric~M.~Wolff}{\{cyrus.huang, eric.wolff\}@getcruise.com}

\icmlkeywords{Machine Learning, ICML}

\vskip 0.3in
]



\printAffiliationsAndNotice{}  

\begin{abstract}
We present~\ourmethod{}, a scalable behavior model for autonomous driving. 
We model driving as a sequential decision-making task, and learn a transformer model to predict future agent states as tokens in an autoregressive fashion. 
We scale up our model parameters and training data by multiple orders of magnitude, enabling us to explore the scaling properties in terms of dataset size, model parameters, and compute. 
We evaluate~\ourmethod{} across different scales in a planning task, through both quantitative metrics and qualitative examples, including closed-loop driving in complex real-world scenarios. 
In a separate prediction task,~\ourmethod{} outperforms state-of-the-art baselines and exhibits improved performance by pretraining on a large-scale dataset, further validating the benefits of data scaling.
\end{abstract}
    
\section{Introduction}
\label{sec:intro}

Transformer-based foundation models have become increasingly prevalent in sequential modeling tasks across various machine learning domains. 
These models are highly effective in handling sequential data by capturing long-range dependencies and temporal relationships. 
Their success has been evident in natural language processing~\cite{mann2020language,kaplan2020scaling,hoffmann2022training}, time-series forecasting~\cite{zhou2021informer}, and speech recognition~\cite{kim2022squeezeformer}, where sequential patterns play a crucial role. 
One of the key strengths of transformer-based models is their capacity to learn from large datasets including millions of training examples, enabling them to address complex tasks with increased model sizes, up to billions of model parameters. 

\begin{figure}[t]
  \centering
    \begin{subfigure}
        \centering
        \includegraphics[width=1\linewidth]{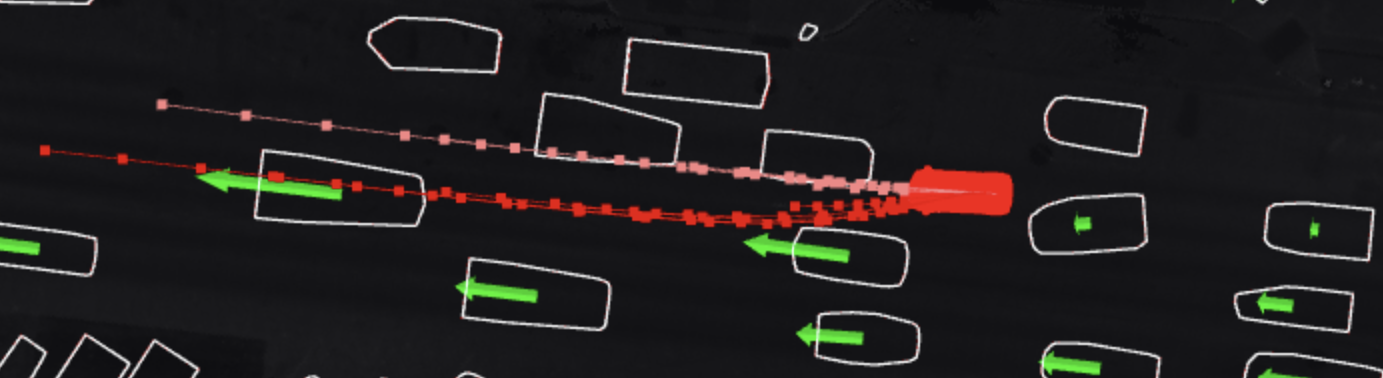}
    \end{subfigure}\vspace{-4mm}\\
    \begin{subfigure}
        \centering
        \includegraphics[width=1\linewidth]{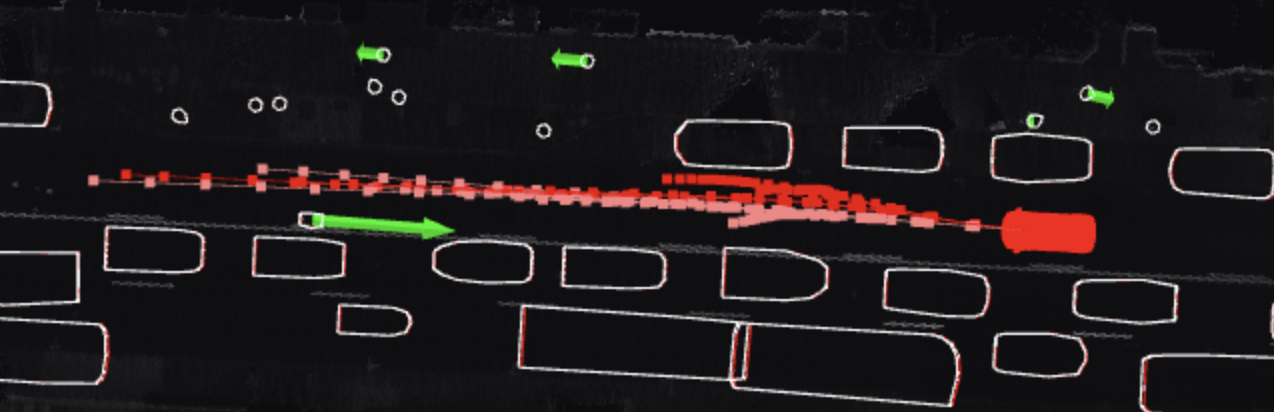}
    \end{subfigure}%
    \vspace{-5mm}
    \caption{Through data and model scaling,~\ourmethod{} (red) handles complex real-world driving scenarios, such as lane changing in heavy traffic and yielding to a cyclist in the opposite lane, compared to a smaller baseline trained on less data (pink).}
    \label{fig:intro}
\end{figure}

\begin{table}[t]
  \centering
  \begin{tabular}{rcccc}
  \toprule
     Model & Parameters & Training Segments \\
     \midrule
    VectorNet & 72K &  211K \\
    PlanTF & 2.1M &  1M \\
    QCNet-Argo2 & 7.3M & 200K \\
    QCNet-WOMD & 7.5M & 2.2M \\
    Wayformer & 20M &  2.2M \\
    MotionLM & 27M & 1.1M \\
    MTR & 66M & 2.2M \\
    GUMP & 523M & 2.6M \\
    \textbf{\ourmethod{} (Ours)} & \textbf{1.4B} & \textbf{120M} \\
    \bottomrule
  \end{tabular}
  \vspace{-2mm}
  \caption{\ourmethod{} is $\sim$3x larger and is trained on $\sim$50x more data sequences than existing published behavior models. 
  }
  \label{tab:model_survey}
  \vspace{-4mm}
\end{table}

While scaling up model and dataset sizes has been critical for recent advances in sequential modeling for text prediction~\cite{kaplan2020scaling,hoffmann2022training}, it remains unclear whether these scaling trends can be directly extended to behavior modeling, particularly in driving tasks, due to several unique challenges.
First, driving tasks involve a wider range of input modalities, including agent trajectories and map information, unlike language tasks that rely solely on textual inputs.
Second, behavior modeling demands spatial reasoning and an understanding of physical kinematics. Such capabilities are typically beyond the scope of language models.
Finally, the collection of large-scale driving datasets requires substantial effort and resources, making it far more challenging than gathering textual data.
As a result, existing work is often constrained by the availability of training data or the scalability of the models, as summarized in Table~\ref{tab:model_survey}.

In our work, we present a comprehensive study of scaling up data sizes and model parameters in the context of behavior modeling for autonomous driving, which predicts future actions of traffic agents to support critical tasks such as planning and motion prediction.
Specifically, we train a transformer-based autoregressive behavior model on over 100 million high-quality human driving examples, $\sim$50 times more than existing open-source datasets, and scale the model over 1 billion parameters, outsizing existing published behavior models. 

As we scale up the volume of training data and the number of model parameters, we observe improvements in both quantitative metrics and qualitative behaviors. 
More importantly, large models trained on extensive, diverse datasets can better handle rare or edge-case scenarios, which often pose significant challenges for autonomous vehicles, as shown in Fig.~\ref{fig:intro}\footnote{The baseline model is trained on $\sim$50 times less data and uses $\sim$50 times fewer parameters.}. 
As a result, we see great potential in scaling up behavior models through data and model parameters to improve the safety and robustness of autonomous driving systems. 

\begin{figure*}
    \centering
    \includegraphics[width=1\linewidth]{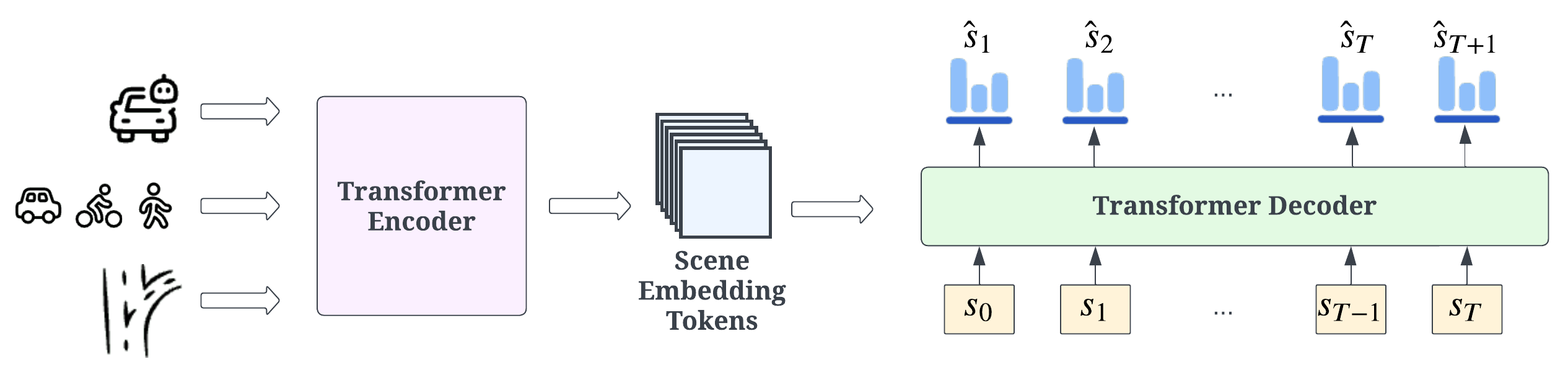}
    \caption{\ourmethod{} architecture, including a transformer encoder and a transformer decoder. The transformer encoder summarizes relevant scene context, such as target agent history, nearby agent history, and map information, into a set of scene embedding tokens. The transformer decoder follows an LLM-style architecture that takes a sequence of agent states as input and predicts a discrete distribution of actions at the next tick, conditioned on previous states.}
    \label{fig:architecture_diagram}
\end{figure*}

Our main contributions are as follows:
\begin{enumerate}
    \item We present~\ourmethod{}, a large autoregressive behavior model for driving, by scaling up both model parameters and real-world training data samples. 
    \item We determine empirical driving scaling laws for an autoregressive behavior model in terms of data size, model parameters, and compute. We validate the value of scaling up training data and compute, and observe better model scalability as training data increases, consistent with the language scaling literature. 
    \item We quantitatively and qualitatively compare models from our scaling experiments to validate their effectiveness in real-world driving scenarios. We present real-world deployment of our model through closed-loop driving in challenging conditions. 
    \item We demonstrate the generalizability of our model on the Waymo Open Motion Dataset, which outperforms prior state-of-the-art on the motion prediction task and achieves improved performance through large-scale pretraining.
\end{enumerate}

\vspace{-4mm}
\section{Related Work}
\label{sec:related_work}

\subsection{Behavior Modeling}
Behavior modeling is a critical task in autonomous driving, which covers a broad spectrum of tasks including planning, prediction, and simulation. 
Taking multimodal inputs including agent history states and map information, behavior models predict the future states of these traffic agents by reasoning about agent dynamics~\cite{cui2020deep,song2022learning}, interactions~\cite{sun2022m2i,jiang2023motiondiffuser}, human intent~\cite{shi2022motion,huang2022hyper,sun2024controlmtr}, and driving environments~\cite{liang2020learning,kim2021lapred}.

Among learning-based models, transformers have gained popularity due to their ability to fuse multimodal inputs as encoders~\cite{nayakanti2023wayformer,zhang2023hptr,zhou2023query,jia2023hdgt,gan2024multi} and model long-range temporal relationships as decoders~\cite{seff2023motionlm,shi2022motion,shi2024mtr++}.
Despite the success of transformers in behavior modeling, existing literature is often restricted by the size of model parameters due to limited training data, which fails to capture the full scaling potential of transformer-based models. 
In our work, we scale up our transformer models to include billions of parameters, by training on a large-scale dataset including more than 100M driving demonstrations, and validate the scalability of transformer-based models in the context of autonomous driving.

\vspace{-2mm}
\subsection{Large Transformer Models}
Large transformers have demonstrated great success in sequential modeling tasks, by scaling up model parameters and data sizes~\cite{kaplan2020scaling,hoffmann2022training,zhai2022scaling,muennighoff2023scaling}.
These scaling laws have pushed the boundary of many sequential modeling tasks including natural language processing~\cite{mann2020language}, time-series forecasting~\cite{zhou2021informer}, and speech recognition~\cite{kim2022squeezeformer}.

Recent work has studied the scalability of behavior models in the context of motion prediction~\cite{seff2023motionlm,ettinger2024scaling}, planning~\cite{sun2023large}, and simulation~\cite{hu2025solving}, yet these studies are either constrained by limited data size (up to a couple of million training examples), or focused on a few orders of magnitude in terms of data and model scaling, limiting the potential to draw statistically significant conclusions over a large scaling range.

In this paper, we study the scaling properties (in terms of data samples, model parameters, and compute) across a much larger range compared to prior work. 
More specifically, we target an autoregressive decoder architecture that has been proven to be both scalable (as in the LLM literature) and effective in generating accurate trajectories for traffic agents~\cite{seff2023motionlm,chen2024drivinggpt,hu2024drivingworld}. 

Beyond autonomous driving, there is limited relevant literature on building large transformer models for robotic tasks~\cite{o2024open,octo_2023}, which share a transformer architecture similar to our work. 
Robotics models often share different input features and dynamic models, and operate in different environments, making them difficult to apply directly to driving tasks.

\vspace{-2mm}
\section{Behavior Model}
\label{sec:model}

We use a standard encoder-decoder architecture as our behavior model, as shown in Fig.~\ref{fig:architecture_diagram}.
We use transformer-based models as our encoder and decoder backbones due to their scalability in related sequential modeling tasks.

\subsection{Problem Formulation}
We model the problem as a sequential prediction task over the future positions of the target agent up to horizon $T$, by applying the chain rule at each step, conditioning on driving context information $\mathbf{c}$ and historical agent positions $\mathbf{s}$:
\begin{equation}
\label{eq:prob}
    P(s_{1:T} | \mathbf{c}) = \Pi_{t=1}^{T} P(s_{t} | s_{0:t-1}, \mathbf{c}).
\end{equation}
The context information includes target agent history states $c_{\text{target}}$, nearby agent history states $c_{\text{nearby}}$, and map states $c_{\text{map}}$. The historical agent information includes agent positions from previous historical steps, i.e.~$s_{0:t-1}$ if we want to predict agent positions at step $t$. 

We define ``state'' as a complete kinematic state including position, orientation, velocity, and acceleration, which is commonly available in agent history observations, and ``position'' as 2-D (x, y) coordinates to simplify the output space.

\subsection{Scene Encoder}
\label{sec:encoder}
The encoder follows a standard transformer encoder architecture~\cite{shi2022motion} that fuses all input modalities into a set of scene embedding tokens. 
It consumes raw input features, including target agent history states, nearby agent history states, and map states as a set of vectors, and normalizes all inputs to an agent-centric view. 
Each vector is mapped to a token embedding through a PointNet-like encoder as in~\cite{gao2020vectornet}. 
At the end of the encoder, we apply a self-attention transformer~\cite{vaswani2017attention} to fuse all input context into a set of encoder embeddings, $\mathbf{c} \in \mathbb{R}^{n \times d}$, where $n$ is the number of vectors and $d$ is the token dimensions, that summarize the driving scene.

\subsection{LLM-Style Trajectory Decoder}
\label{sec:decoder}
Inspired by the LLM literature~\cite{radford2019language}, we follow~\cite{seff2023motionlm} to use a transformer decoder architecture to predict the distribution of agent positions at each step in the future. 

The decoder first tokenizes agent positions at all steps into embeddings with dimension $d$ through a linear layer, followed by a LayerNorm layer and a ReLU layer.
At each step $t$, the decoder takes agent embeddings up to $t$, and cross attends them with the encoder embeddings $\mathbf{c}$ to predict the distribution of agent positions at the next step $t+1$.

The output is a set of discrete actions $a$ represented as the Verlet action~\cite{rhinehart2018r2p2,seff2023motionlm}, as the second derivative of positions. We can apply the following equation to map Verlet actions to positions:
\begin{equation}
    s_{t+1} = s_{t} + (s_{t} - s_{t-1}) + a_{t},
\end{equation}
where $a_{t}$ is the predicted Verlet action, and $(s_{t} - s_{t-1})$ assumes a constant velocity step. 
This representation helps predict smooth trajectories using a small set of actions. 

\subsection{Training}
To train a~\ourmethod{} model, we follow teacher forcing by applying ground truth future positions as input to the trajectory decoder. 
This allows us to predict all future steps in parallel.

We use a single cross-entropy classification loss over the action space, where the target action is selected as the one that is closest to the ground truth future trajectories. We refer the reader to Appendix~\ref{sec:training_detail} for more training details.

\subsection{Inference}
\label{subsec:inference}
At inference time, we follow a standard LLM setup and roll out a trajectory over horizon $T$ autoregressively, by repeating the process of predicting the action distribution at the next step, sampling an action, and adding it back to the input sequence.

We sample multiple trajectories in batch to approximate the distribution and then subsample to the desired number of modes using $K$-Means, as in~\cite{seff2023motionlm}.

\section{Scaling Experiments}
\label{sec:scaling}

The goal of our scaling experiments is to determine the effect of data and model size on behavior prediction performance.
Quantifying scaling laws similar to those seen in LLMs~\cite{kaplan2020scaling} can help prioritize the value of data and compute for future research directions in behavior modeling.
We focus our effort on exploring the next frontier of data and model size -- over an order of magnitude beyond previously published work.

\paragraph{Large-scale driving dataset} From millions of miles of high-quality real-world human driving demonstrations, we curate a small subset of 120M segments for an internal research dataset. 
The dataset is carefully curated and balanced to represent diverse geographic regions across multiple cities and countries, including the United States, Japan, and the UAE. Data collection is evenly distributed between daytime and nighttime and is conducted primarily in urban environments. The dataset captures a wide range of challenging driving scenarios, such as lane changes, intersections, double-parked vehicles, construction zones, and close interactions with pedestrians and cyclists.

We extracted map information, target agent states, and nearby agent states into vectorized representation, as customary in behavior modeling literature~\cite{gao2020vectornet}. 

\paragraph{Scaling overview}
We scale the model size across three orders of magnitude, from 1.5 million to 1.4 billion parameters, by increasing the embedding dimension in both encoder and decoder transformers. 
For each model size, we explore multiple learning rate schedules using a cosine decay over the full training steps and select the maximum learning rate that yields the best performance. 
Table.~\ref{tab:model_scaling_lr} summarizes the optimal learning rate for each model size.
Consistent with practices in large language model scaling~\cite{kaplan2020scaling}, each model is trained for a single epoch.

\begin{table}[t]
  \centering
  \setlength{\tabcolsep}{3pt}
  \begin{tabular}{r|cl}
  \toprule
     Model Size & Hidden Dimension~($d_{\text{model}}$) & Max LR \\
     \midrule
     1.5M & 64 & 0.0050 \\
     4M & 128 & 0.0020 \\
     8M & 192 & 0.0014 \\
     12M & 256 & 0.0010 \\
     26M & 384 & 0.0010 \\
     94M & 768 & 0.0007 \\
     163M & 1024 & 0.0004 \\
     358M & 1536 & 0.0002 \\
     629M & 2048 & 0.0001 \\
     1.4B & 3072 & 0.0001 \\
    \bottomrule
  \end{tabular}
  \caption{We vary the model size over three orders of magnitude through hidden dimensions. For each model size, we report the optimal learning rate, which decreases as the model size increases, matching the observations in the LLM scaling literature~\cite{kaplan2020scaling,hoffmann2022training}.}
  \label{tab:model_scaling_lr}
\end{table}

We evaluate model performance using validation loss, computed on a comprehensive validation set of 10 million samples drawn from the same distribution as the training data, with no overlap.
This set remains fixed across all scaling experiments to ensure consistency.
We use validation loss as a proxy to measure model performance, following standard practices in scaling studies~\cite{kaplan2020scaling,hoffmann2022training,muennighoff2023scaling}.
This loss, calculated as cross-entropy on next-action prediction, serves as our primary performance metric. 
Additional driving-specific metrics are reported in Sec.~\ref{sec:av_planning}.

\subsection{Data Scaling}
\label{sec:scaling_data}


\begin{figure}[t]
    \centering
    \includegraphics[width=0.9\linewidth]{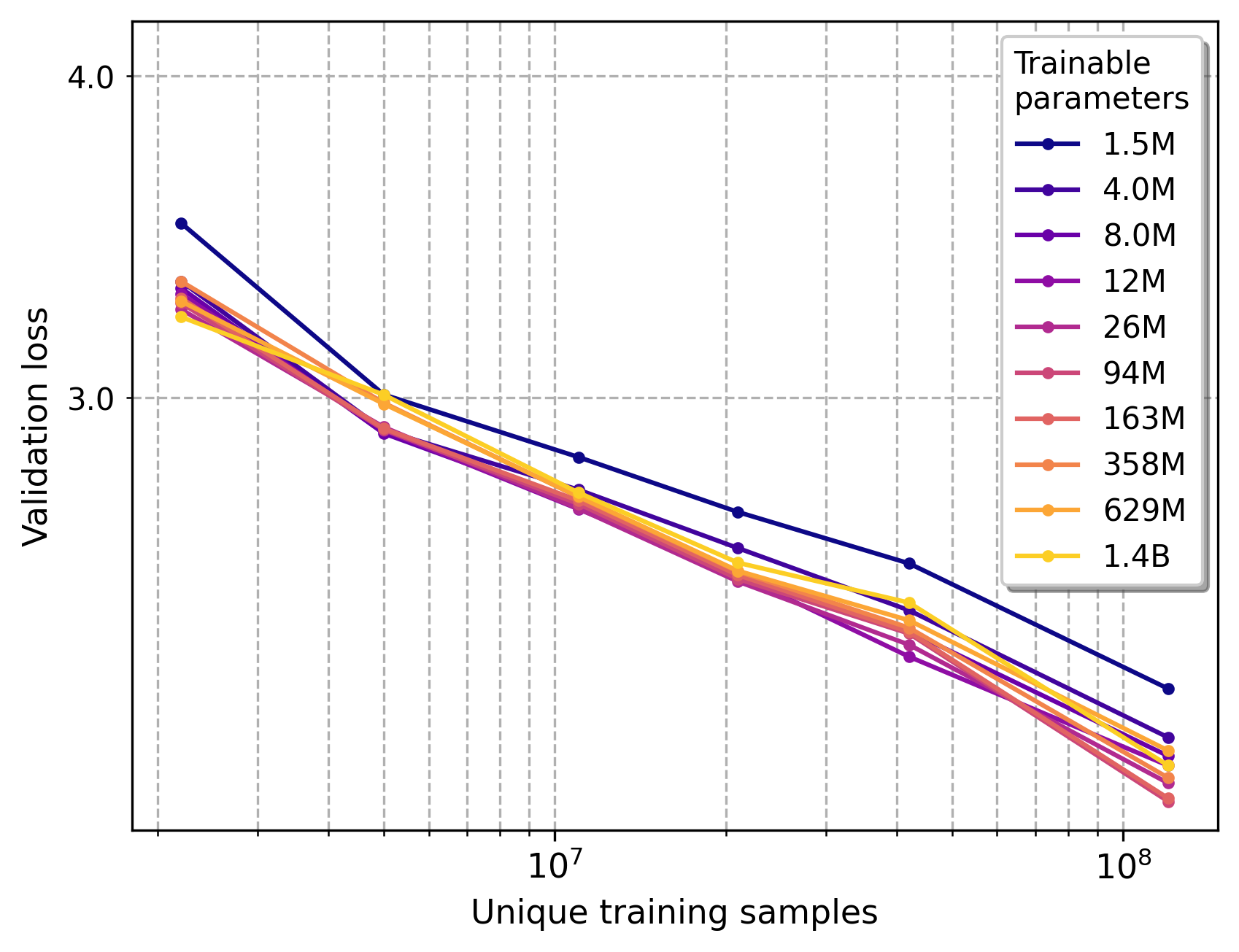}
    \caption{Performance increases with dataset size across a range of model parameters, indicating that data is a limiting factor. Both axes are on a log scale. An exponential fit was applied to all data points except for the 1.5M curve, resulting in the following relationship: 
   $\log(L) = -0.102 \log(D) + 2.663$ with an $R^2$ value of 0.986, where $L$ is the validation loss and $D$ is the number of unique training samples.} 
    \label{fig:scaling_data}
\end{figure}

Data scaling results are summarized in Fig.~\ref{fig:scaling_data}. 
The smallest dataset of 2.2M samples mimics the size of Waymo Open Motion Dataset (WOMD)~\cite{ettinger2021large}, a large open-source dataset for behavior modeling ($\sim$44k scenarios with multiple target agents per scenario). 
We select a few subsets of our internal research dataset to study data scaling across different orders of dataset sizes.
Our experiments use $\sim$50x more data than WOMD, exploring a new region of the design space.

The results indicate that as the model is trained on more unique data samples, the performance improves, regardless of model size.
 Extrapolating from the scaling law in Fig.~\ref{fig:scaling_data}, to improve the best loss by another $10\%$, we would need to include $350$M more training examples.
A $20\%$ improvement would require about $1.4$B more examples. As a result, we find that data remains the bottleneck for further improving driving performance.

Lastly, the scaling results remain relatively consistent across model sizes.
This consistency indicates that data scaling comparisons can be done on reasonably small model sizes beyond 10M parameters.

\subsection{Model Scaling}
\label{sec:scaling_model}

\begin{figure}[t]
    \centering
    \includegraphics[width=1\linewidth]{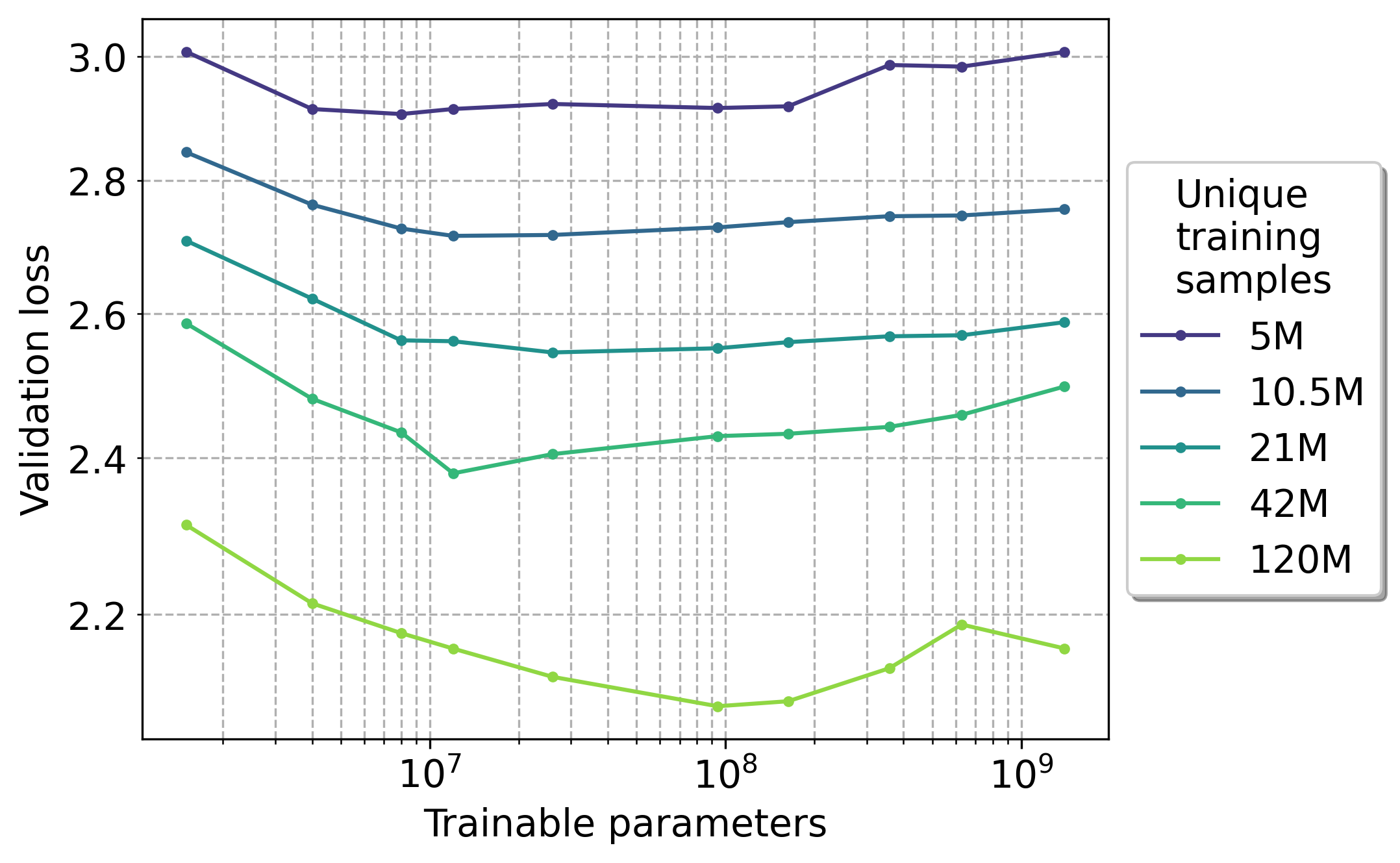}
    \caption{Model scaling is more effective as training data increases. The validation loss improves up to~$\sim$100M parameters when trained on the full dataset.}
    \label{fig:scaling_model}
\end{figure}

We now study model sizes across three orders of magnitude (1.5M to 1.4B parameters), as listed in Table~\ref{tab:model_scaling_lr}.
We increase model size by increasing the hidden dimensions of transformers for simplicity. 
We notice that modifying other parameters such as number of attention heads and hidden dimensions per head does not lead to noticeable changes in the results, as studied in Appendix~\ref{sec:scaling_detail}. 

Training larger models is sensitive to learning rates, as observed in other scaling studies~\cite{kaplan2020scaling,hoffmann2022training}. 
For each model size, we run multiple experiments at different learning rates to select the one with the optimal performance, as summarized in Table.~\ref{tab:model_scaling_lr}.

Results in Fig.~\ref{fig:scaling_model} demonstrate that increasing the amount of training data enhances the effectiveness of model scaling. 
Specifically, when the dataset size is up to 21M, the impact on the validation loss is barely noticeable across a large range of model sizes. 
Beyond 21M samples, validation losses improve with larger models -- up to 12M parameters when trained on the 42M dataset and up to 94M parameters when trained on the 120M dataset -- before reaching a plateau and eventually overfitting. 
These findings further reinforce that data is the primary bottleneck for scaling models, aligning with observations in the LLM scaling literature~\cite{kaplan2020scaling}. 
We leave further exploration of scaling behavior with larger datasets to future work.

\begin{figure}[t]
    \centering
    \includegraphics[width=\linewidth]{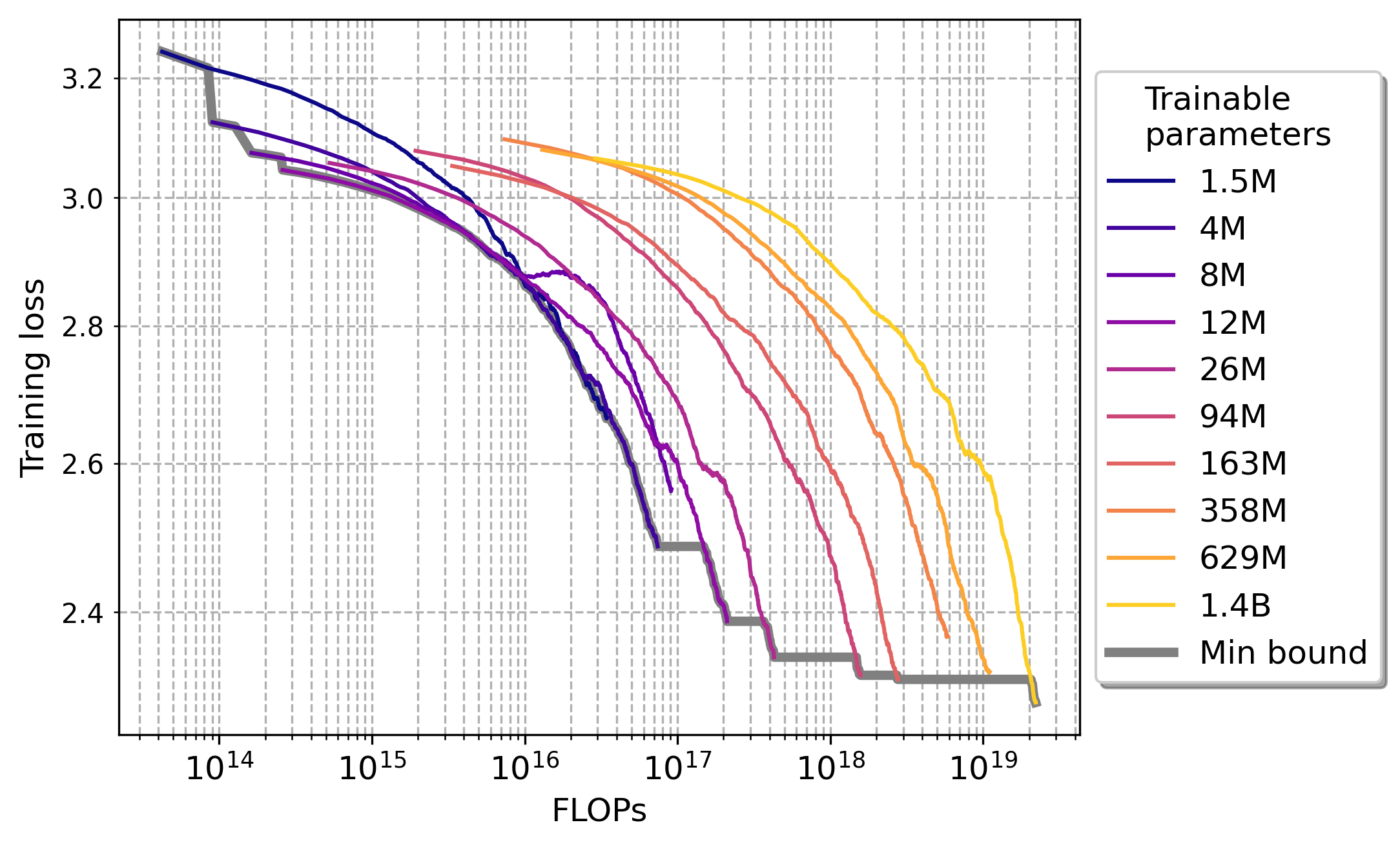}
    \caption{Relationship between (smoothed) training loss and FLOPs. Each curve represents an experiment corresponding to a specific model size, and the ``min bound'' indicates the best performance possible for a given FLOP budget.}
    \label{fig:scaling-training-curves}
\end{figure}

\begin{figure}[t]
    \centering
    \includegraphics[width=\linewidth]{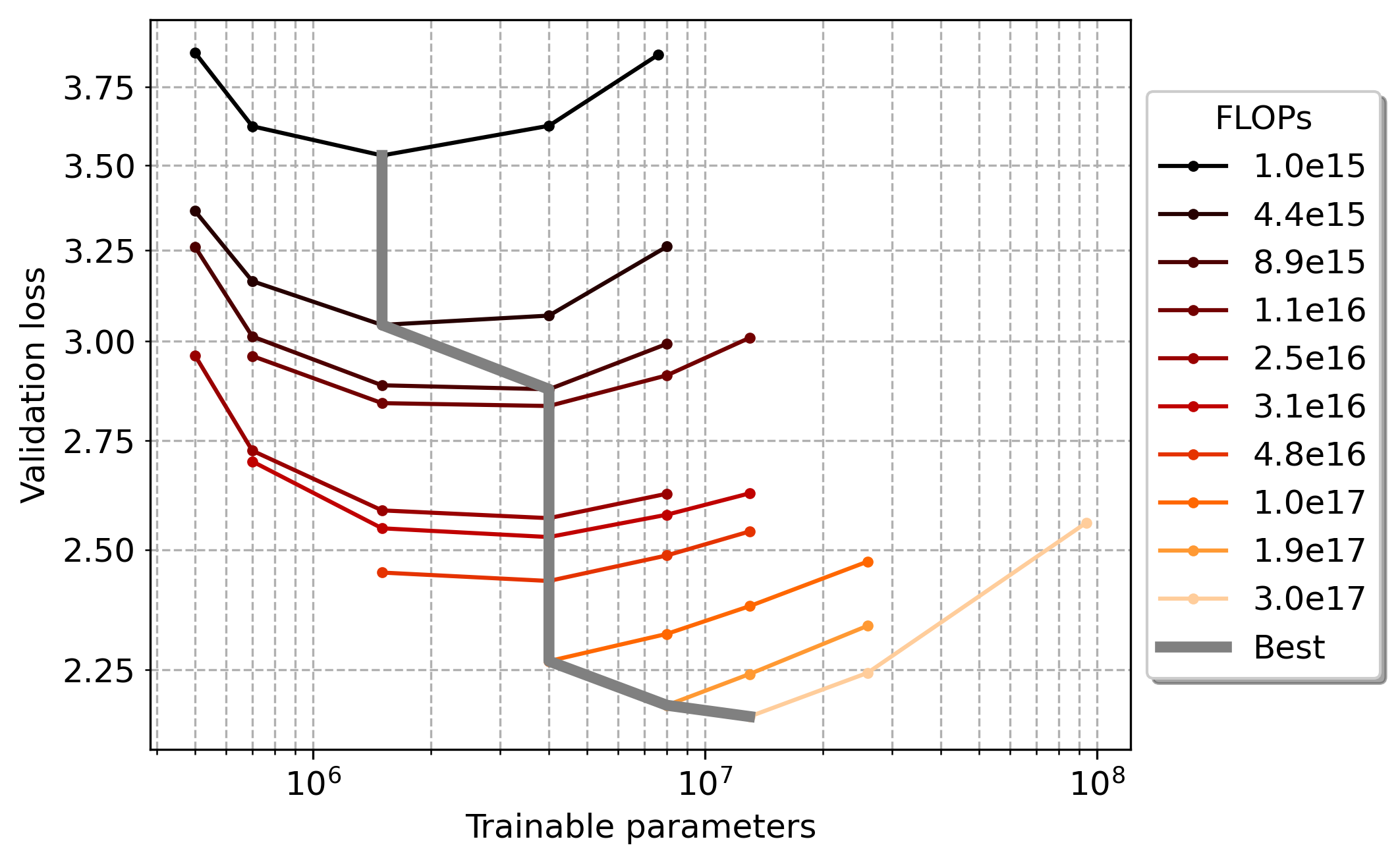}
    \vspace{-4mm}
    \caption{Performance as a function of model size for fixed compute budgets. The solid gray line connects the best result from each compute budget.}
    \label{fig:iso-flops-loss}
\end{figure}

\subsection{Compute Scaling}
In Fig.~\ref{fig:scaling-training-curves}, we examine how compute affects training loss, where compute is measured by Floating Point Operations (FLOPs).
We identify a monotonically decreasing ``min-bound'' boundary, which shows the lowest training loss observed up to the current compute value. 
As we increase compute, training loss generally decreases. 
Initially, this decrease is quite steep, but it gradually slows down at higher FLOPs values. 
This trend is consistent with observations in the LLM scaling literature, such as those reported by~\cite{hoffmann2022training}, covering a subset of the full FLOP range explored in these studies.

Next, we investigate whether there is an optimal combination of data size and model parameters for a fixed compute budget.
Given the large computational expense for training models at the scales we are exploring, it is important to make the best use of our data.
In this study, we fixed the compute budget in different FLOP groups.
For a fixed compute budget, we can allocate resources either to model parameters or data samples, keeping their product constant.

Fig.~\ref{fig:iso-flops-loss} plots the performance with different compute budgets.
The trend clearly shows that a larger compute budget leads to better performance, with optimal model size increasing accordingly, as indicated by the ``best" gray line. 
The results further reveal that data is the main bottleneck, as the smallest model outperforms others in the three largest FLOP groups. 

\subsection{Ablation Study on Decoder Architecture}
\label{sec:scaling_ablation}

We scale up two different model architectures by two orders of magnitude: our autoregressive decoder and a one-shot decoder. For the one-shot decoder, we follow~\cite{nayakanti2023wayformer} to use a transformer decoder that takes a set of learned queries and cross attends them with scene embeddings to produce trajectory samples. This decoder is referred to as ``one-shot'' because it generates the full trajectory rollout at once, where an autoregressive decoder follows an LLM-style to produce trajectories one step at a time.

\begin{figure}[t]
    \centering
    \includegraphics[width=0.9\linewidth]{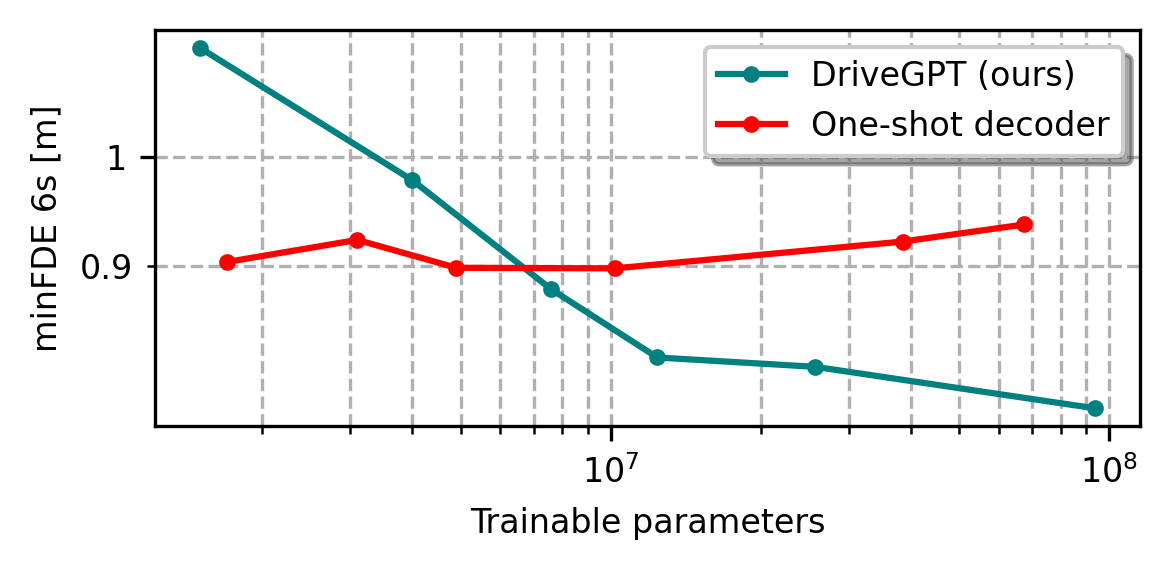}
    \vspace{-4mm}
    \caption{~\ourmethod{} achieves better scalability in terms of minFDE, by adopting an autoregressive architecture compared to a one-shot architecture.}
    \label{fig:ablation_both}
    \vspace{-4mm}
\end{figure}

The results are summarized in Fig.~\ref{fig:ablation_both}, where we use minFDE at 6 seconds as a proxy to measure the model performance because of different loss definitions between two decoder architectures. Despite worse performance at small-scale parameters, our autoregressive decoder achieves better scalability and outperforms the one-shot baseline beyond 8M parameters. While we find it harder to scale the one-shot decoder, we confirm that our autoregressive decoder scales up to 100M parameters in terms of prediction accuracy, and defer further scalability study on the one-shot decoder as future work.

\section{Planning and Prediction Experiments}
\label{sec:experiments}

In this section, we show detailed results of~\ourmethod{} in a planning task using our internal research dataset and a motion prediction task using an external dataset.
The results here further explore the impact of scaling from Section~\ref{sec:scaling} and help ground those results in driving tasks and metrics.

\subsection{Internal Evaluation: AV Planning}
\label{sec:av_planning}
For the planning task, we train our model using our internal research dataset, composed of millions of high-quality human driving demonstrations, and generate AV trajectories by autoregressively predicting the next action, as described in Sec.~\ref{subsec:inference}. 

We approximate the distribution by oversampling trajectories in batch and subsampling to 6 trajectories as in~\cite{seff2023motionlm}.
While the AV must ultimately select a single trajectory for planning, returning multiple samples helps better understand multimodal behavior and aligns with motion prediction metrics.

We measure the planning performance on a comprehensive test set through a set of standard geometric metrics including minADE (mADE), minFDE (mFDE), and miss rate (MR).
Additionally, we use semantic-based metrics including offroad rate (Offroad) that measures the ratio of trajectories that leave the road and collision rate (Collision) that measures the ratio of trajectories overlapping with traffic agents.
We normalize these metrics across experiments to highlight relative performance changes.

\begin{table}[t]
  \centering
  \setlength{\tabcolsep}{3pt}
  \begin{tabular}{r|ccccc}
  \toprule
     Data & $\widehat{\textrm{mADE}}\downarrow$ & $\widehat{\textrm{mFDE}}\downarrow$ & $\widehat{\textrm{MR}}\downarrow$ & $\widehat{\textrm{Offroad}}\downarrow$ & $\widehat{\textrm{Collision}}\downarrow$ \\
     \midrule
    2.2M & 1.000 & 1.000 & 1.000 & 1.000 & 1.000 \\
    21M & 0.561 & 0.496 & 0.420 & 0.326 & 0.269 \\
    85M & 0.496 & 0.441 & 0.332 & 0.238 & 0.217 \\
    120M & \bf{0.489} & \bf{0.433} & \bf{0.317} & \bf{0.198} & \bf{0.196} \\
    \bottomrule
  \end{tabular}
      \caption{As we scale up more training data,~\ourmethod{} produces better AV trajectories as measured across all metrics. Metrics are normalized to highlight relative performance.}
  \label{tab:data_scaling}
\end{table}

\begin{figure}[t]
  \centering
    \begin{subfigure}
        \centering
        \includegraphics[width=1\linewidth,trim=0 5pt 0 25pt,clip]{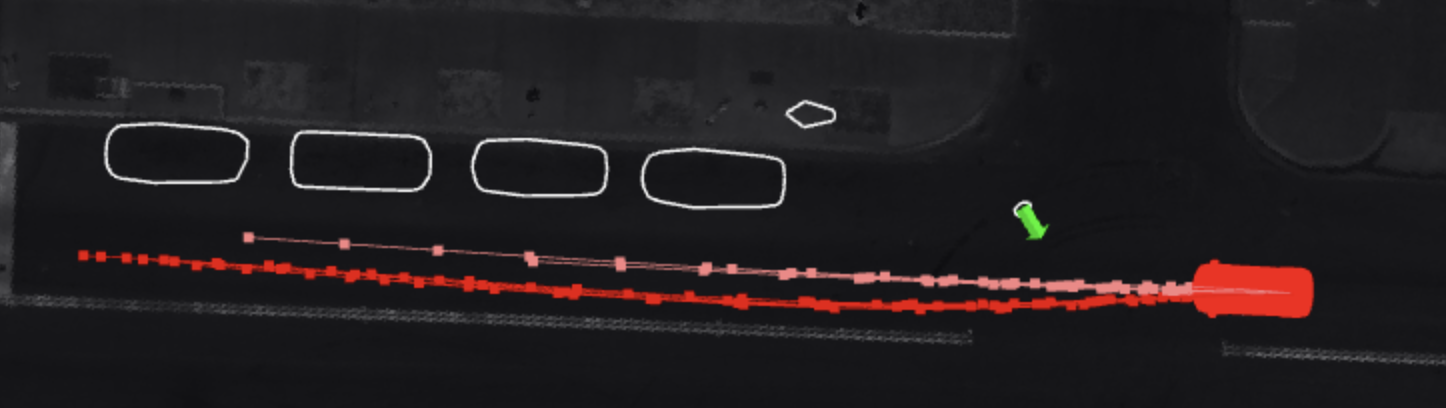}
    \end{subfigure} \vspace{-4mm}\\
    \begin{subfigure}
        \centering
        \includegraphics[width=1\linewidth]{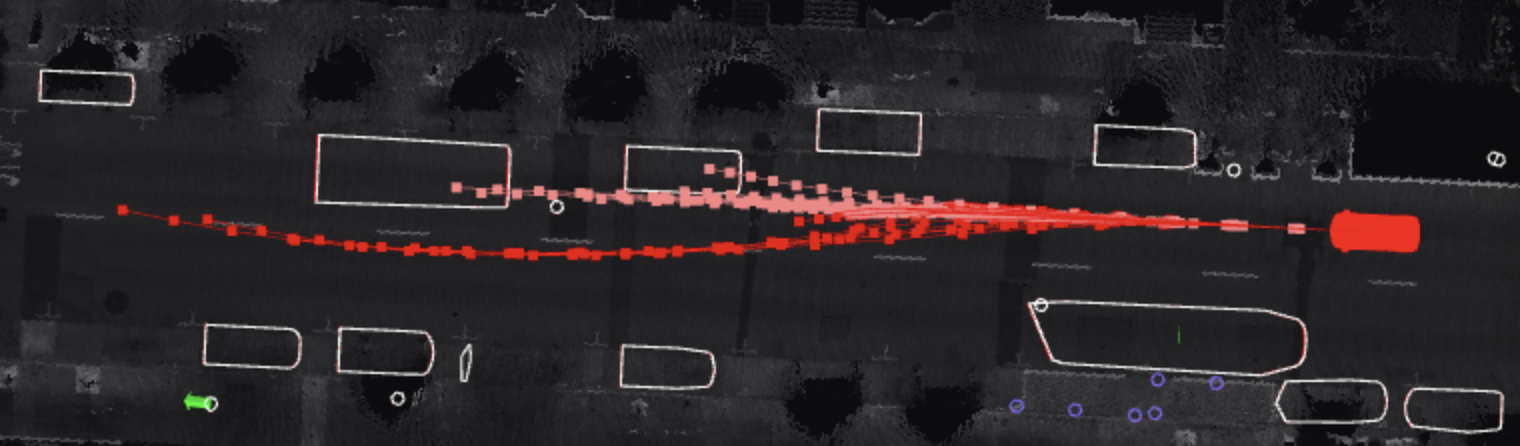}
    \end{subfigure}%
    \caption{By training on 50x more data,~\ourmethod{} (red) is able to produce high-quality trajectories in red that keep a safe lateral distance from a jaywalker (top) and go around two double-parked vehicles (bottom).}
    \label{fig:data_scaling_qualitative}
    \vspace{-4mm}
\end{figure}

\subsubsection{Data Scaling Results}
\label{exp:data_scaling}

We compare a~\ourmethod{} model at 26M parameters trained on datasets of different sizes. The baseline dataset (2.2M) is selected to mimic the size of a typical behavior modeling dataset such as WOMD. 

The results are presented in Table~\ref{tab:data_scaling}, where we see that training on more data samples significantly improves the quality of the predicted AV trajectories, in terms of critical semantics metrics in driving including offroad rate and collision rate, as well as geometric metrics.
These improvements are consistent with Sec.~\ref{sec:scaling_data}. 

We further present two qualitative examples in Fig.~\ref{fig:data_scaling_qualitative} to illustrate the value of training on more data. In these examples, red trajectories represent~\ourmethod{} trained on 120M samples, and pink trajectories are from the same model trained on 2.2M samples. The examples show that our method produces map-compliant and collision-free trajectories when trained on more data, successfully handling complicated interactions involving a jaywalking pedestrian and two double-parked vehicles.

\subsubsection{Model Scaling Results}
\label{exp:model_scaling}

\begin{table}[t]
  \centering
  \setlength{\tabcolsep}{3pt}
  \begin{tabular}{r|ccccc}
  \toprule
     Model & $\widehat{\textrm{mADE}}\downarrow$ & $\widehat{\textrm{mFDE}}\downarrow$ & $\widehat{\textrm{MR}}\downarrow$ & $\widehat{\textrm{Offroad}}\downarrow$ & $\widehat{\textrm{Collision}}\downarrow$ \\
     \midrule
    8M & 1.000 & 1.000 & 1.000 & 1.000 & 1.000 \\
    26M & 0.954 & 0.950 & 0.902 & 0.858 & 0.915 \\
    94M & \bf{0.937} & \bf{0.925} & \bf{0.866} & \bf{0.815} & 0.890 \\
    163M & 0.943 & \bf{0.925} & 0.875 & \bf{0.815} & \bf{0.817} \\
    \bottomrule
  \end{tabular}
  \caption{As we scale up more model parameters, all metrics improve up to 94M. Collision rate continues to improve at 163M. Metrics are normalized to highlight relative performance.}
  \label{tab:model_scaling}
  \vspace{-1mm}
\end{table}

\begin{figure}[t]
  \centering
    \begin{subfigure}
        \centering
        \includegraphics[width=1\linewidth, trim=0 15pt 0 25pt,clip]{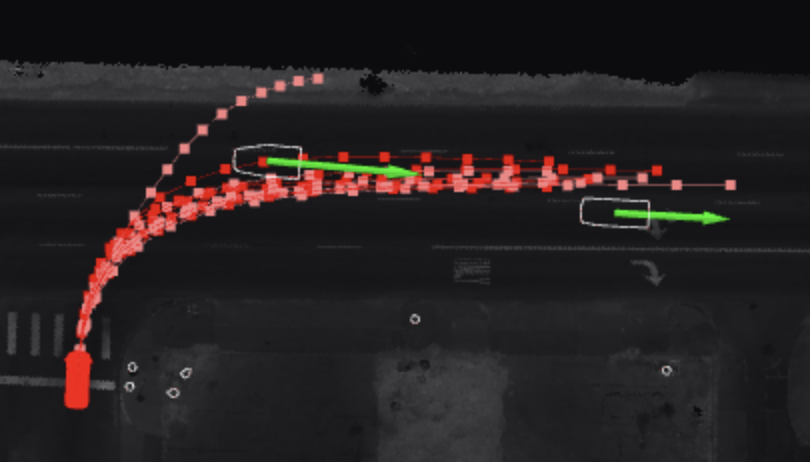}
    \end{subfigure}
    \caption{By training with 12x more parameters,~\ourmethod{} (red) produces realistic trajectories that obey lane boundaries when performing a right turn into traffic.}
    \label{fig:model_scaling_qualitative}
\end{figure}

We train four models using our 120M internal research dataset, and select the 8M model as the baseline. The baseline represents a reasonable size at which our model starts to outperform one-shot decoders, as shown in Sec.~\ref{sec:scaling_ablation}.

The results are presented in Table~\ref{tab:model_scaling}, where all metrics improve as the model size increases up to 94M parameters, with further gains in the collision metric at 163M parameters (see Sec.~\ref{sec:extra_qualitative_large_models} for qualitative examples). Although validation loss shows diminishing returns beyond 94M in Fig.~\ref{fig:scaling_model}, driving metrics continue to improve with increased model capacities, highlighting the potential benefits of introducing more parameters for enhanced driving performance.

We present a qualitative example in Fig.~\ref{fig:model_scaling_qualitative}, where a larger~\ourmethod{} including 94M parameters produces better trajectory samples in red that stay within the road boundary, compared to a smaller version including 8M parameters, in a right turn scenario.

\subsubsection{Closed-Loop Driving}
\label{exp:closed_loop_driving}

We demonstrate the effectiveness of~\ourmethod{} as a real-time motion planner deployed in a closed-loop setting. 
The model takes input features from an industry-level perception system that outputs agent states and map information. 
We used the 8M~\ourmethod{} model, trained on the full dataset, to drive the car, achieving a latency of under 50ms on a single onboard GPU.

\begin{figure}[t]
  \centering
    \begin{subfigure}
        \centering
        \includegraphics[width=1\linewidth]{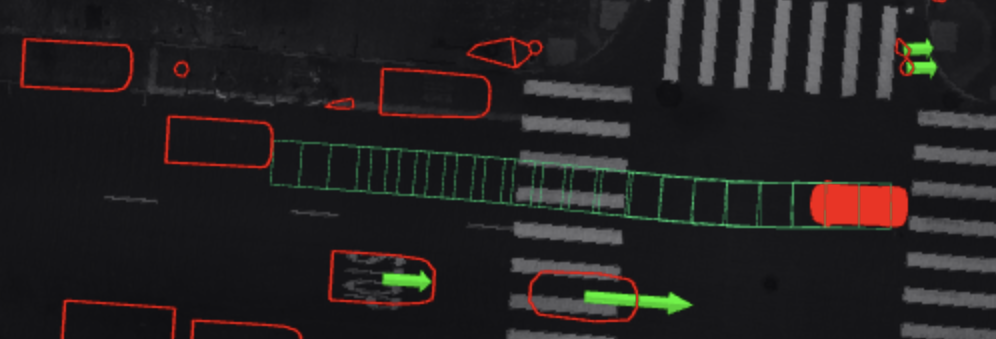}
    \end{subfigure}\vspace{-4mm}\\
    \begin{subfigure}
        \centering
        \includegraphics[width=1\linewidth]{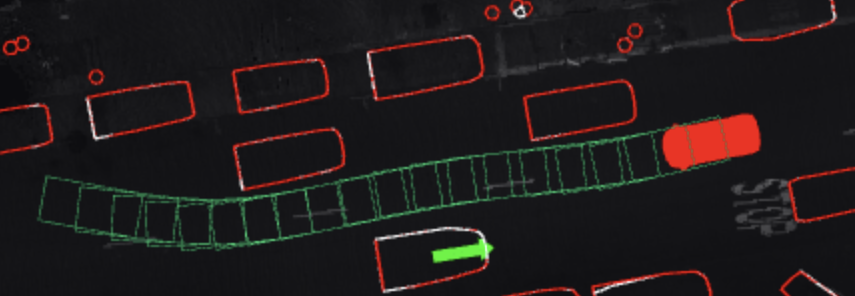}
    \end{subfigure}\vspace{-4mm}\\
    \begin{subfigure}
        \centering
        \includegraphics[width=1\linewidth]{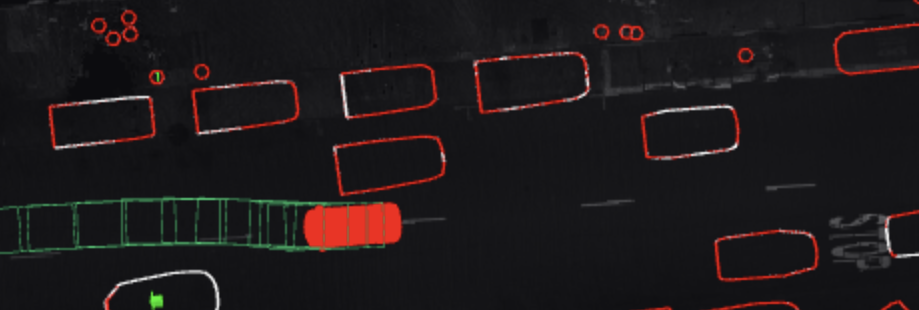}
    \end{subfigure}%
    \vspace{-3mm}
    \caption{Example of~\ourmethod{} running as a closed-loop planner in real time in dense urban traffic. It first produces a forward trajectory (top), updates it to bypass double-parked vehicles (middle), and drives back to the original lane at the end (bottom).}
    \label{fig:closed_loop}
\end{figure}

In Fig.~\ref{fig:closed_loop}, we present a challenging example in dense urban traffic where there are two double-parked vehicles blocking the path forward along with other oncoming vehicles.
~\ourmethod{} generates smooth and safe trajectories, bypassing the blocking vehicles and moving back to the original lane afterward. More examples are presented in the supplementary video at \url{https://youtu.be/-hLi44PfY8g}, where DriveGPT alone is responsible for driving in real time.

\subsection{External Evaluation: Motion Prediction}
\label{sec:external_eval}
To directly compare with published results, we evaluate~\ourmethod{} on the WOMD motion prediction task.
Additionally, we explore the benefits of scale by pretraining on our internal research dataset and finetuning on the significantly smaller WOMD dataset.

\begin{table*}[t]
  \centering
  \setlength{\tabcolsep}{3pt}
  \begin{tabular}{l|ccccc}
  \toprule
      & minADE$\downarrow$ & minFDE$\downarrow$ & Miss Rate$\downarrow$ & Soft mAP$\uparrow$ \\
     \midrule
    MTR~\cite{shi2022motion} & 0.6050 & 1.2207 & 0.1351 & 0.4216 \\
    HDGT~\cite{jia2023hdgt} & 0.7676 & 1.1077 & 0.1325 & 0.3709 \\
    HPTR~\cite{zhang2023hptr} &  0.5565 & 1.1393 & 0.1434 &  0.3968 \\
    ControlMTR~\cite{sun2024controlmtr} & 0.5897 & 1.1916 & 0.1282 & \textbf{0.4572} \\
    MTR++~\cite{shi2024mtr++} & 0.5906 & 1.1939 & 0.1298 & 0.4414 \\
    Wayformer$^{\dagger}$~\cite{nayakanti2023wayformer} & 0.5454 & 1.1280 & 0.1228 & 0.4335 \\
    MotionLM$^{\dagger}$~\cite{seff2023motionlm} & 0.5509 & 1.1199 & \textbf{0.1058} & \underline{0.4507} \\
    \midrule
    \ourmethod{}-WOMD & \underline{0.5279} & \underline{1.0609} & 0.1236 & 0.3795 \\
    \ourmethod{}-Finetune & \textbf{0.5240} & \textbf{1.0538} & \underline{0.1202} & 0.3857 \\
    \bottomrule
  \end{tabular}
  \caption{On the WOMD test set, ~\ourmethod{} achieves better results in geometric metrics without any ensembling, and improved performance after pretraining. Results are averaged over three agent types (vehicle, pedestrian, cyclist) and three prediction horizons (3s, 5s, 8s). The best metric is highlighted in bold and the second best is underlined. $\dagger$ denotes ensemble.}
  \label{tab:womd_test}
\end{table*}

\begin{table}[t]
  \centering
  \setlength{\tabcolsep}{3pt}
  \begin{tabular}{l|cccc}
  \toprule
     Vehicle & minADE$\downarrow$ & minFDE$\downarrow$ & Miss Rate$\downarrow$ \\
     \midrule
    MTR & 0.7642 & 1.5257 & 0.1514 \\
    \ourmethod{}-WOMD & 0.6396 & 1.2823 & 0.1213 \\
    \ourmethod{}-Finetune & \textbf{0.6326} & \textbf{1.2636} & \textbf{0.1181} \\
  \end{tabular}
  \begin{tabular}{l|cccc}
  \toprule
     Cyclist & minADE$\downarrow$ & minFDE$\downarrow$ & Miss Rate$\downarrow$ \\
     \midrule
    MTR & 0.7022 & 1.4093 & 0.1786 \\
    \ourmethod{}-WOMD & 0.6317 & \textbf{1.2617} & 0.1756 \\
    \ourmethod{}-Finetune & \textbf{0.6294} & 1.2641 & \textbf{0.1691} \\
  \end{tabular}
  \begin{tabular}{l|cccc}
  \toprule
     Pedestrian & minADE$\downarrow$ & minFDE$\downarrow$ & Miss Rate$\downarrow$ \\
     \midrule
    MTR & 0.3486 & 0.7270 & 0.0753 \\
    \ourmethod{}-WOMD & 0.3122 & 0.6388 & 0.0738 \\
    \ourmethod{}-Finetune & \textbf{0.3100} & \textbf{0.6338} & \textbf{0.0733} \\
    \bottomrule
  \end{tabular}
  \caption{~\ourmethod{} achieves better geometric metrics across a diverse set of agent types. Results are averaged over 3 prediction horizons.}
  \label{tab:womd_others_all}
  \vspace{-2mm}
\end{table}

\subsubsection{Open-Source Encoder}
For our external evaluation, we use the open-source MTR~\cite{shi2022motion} encoder.
This encoder is similar to the one described in Sec.~\ref{sec:encoder}.
We make this change to improve the reproducibility of our results and take advantage of MTR's open-source dataloading code for WOMD.
We use the same autoregressive decoder as described in Sec.~\ref{sec:decoder}.

\subsubsection{Pretraining Setup}
We made a couple of minor modifications to~\ourmethod{} to be compatible with the WOMD dataset. 
First, we modify our map data to include the same semantics as in WOMD.
Second, we modify our agent data to include the same kinematic features for traffic agents as in WOMD.

We pretrain~\ourmethod{} by training on our internal research dataset for one epoch (as in Sec.~\ref{sec:scaling}).
We load the pretrained checkpoint and finetune the model using the same training setup as in the MTR codebase, where we train the model for 30 epochs using a weighted decay learning rate scheduler.

\subsubsection{Results}
\label{sec:womd_results}
We measure model performance via a set of standard WOMD metrics, including minADE, minFDE, miss rate, and soft mAP. 
Each metric is measured on the test set and computed over three different time horizons.

We present two variants of our method to validate its effectiveness on the motion prediction task, including \textbf{\ourmethod{}-WOMD} that is trained on WOMD, and \textbf{\ourmethod{}-Finetune} that is pretrained on our 120M internal research dataset and finetuned on WOMD. For baselines, we use a set of representative state-of-the-art models.

We report results on the WOMD test set\footnote{Metrics are sourced from~\cite{zhang2023hptr} and WOMD leaderboard.} in Table~\ref{tab:womd_test}. 
The results demonstrate that our method outperforms existing state-of-the-art non-ensemble models in terms of geometric metrics. 
Compared to Wayformer~\cite{nayakanti2023wayformer} and MotionLM~\cite{seff2023motionlm} that use ensembles of up to 8 replicas, our model achieves the best minADE and minFDE metrics and the second-best miss rate metric without any ensembling. 

While we prioritize geometric metrics that emphasize the recall of predicted trajectory samples (i.e. not missing the critical trajectory), our model shows lower soft mAP scores due to suboptimal probability estimates. 
These estimates suffer from accumulated noises over time, as they are computed by compounding the action probabilities over a long sequence (as described in Eq.~\eqref{eq:prob}), resulting in less accurate sample probabilities and reduced soft mAP, which relies on accurate probability assignments across predicted samples.
Consequently, we notice that the gap in soft mAP grows as the prediction horizon increases.
This reveals a limitation of using an autoregressive decoder for accurate probability estimation, as also noted in the LLM literature~\cite{jiang2021can,geng2024survey}. 
We defer improving the probability estimates of autoregressive models for behavior modeling as future work. 
One potential direction is to train an additional probability prediction head for each sample, which could enhance probability estimates and lead to improved soft mAP scores. 

We observe up to 3\% additional gains by pretraining on our internal dataset, despite a large distribution shift between our internal dataset and the public WOMD dataset, in terms of trajectory distributions, feature noises, and differences in semantic definitions.

Diving deeper into agent-specific results in Table~\ref{tab:womd_others_all}, we see consistent improvements across all agent types, compared to MTR that shares the same encoder as ours. This further validates the generalizability of our method, in addition to vehicle behavior modeling results described in Sec.~\ref{sec:av_planning}.

\subsubsection{Qualitative Comparison} 
\begin{figure}[t]
  \centering
    \begin{subfigure}
        \centering
        \includegraphics[width=1\linewidth]{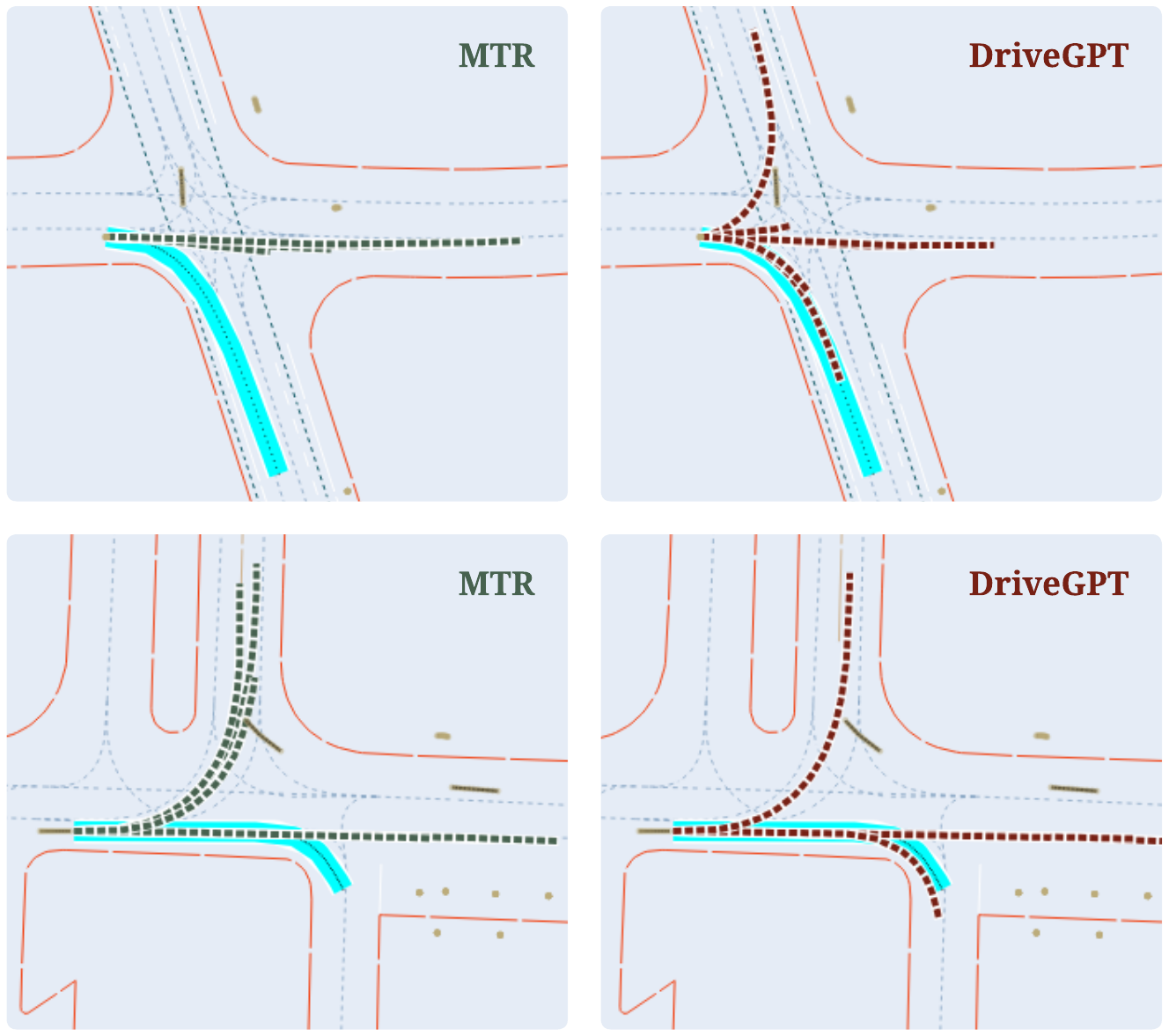}
    \end{subfigure}
    \caption{Compared to MTR (left),~\ourmethod{} (right) produces more accurate and diverse trajectories in complex intersections. Blue represents ground truth future trajectories.}
    \label{fig:womd_qualitative}
    \vspace{-2mm}
\end{figure}

We present two qualitative comparisons in Fig.~\ref{fig:womd_qualitative}, where~\ourmethod{} produces better trajectories in terms of diversity (covering more distinct outcomes) and accuracy (matching with the ground truth future) compared to MTR. This improvement is evident in challenging scenarios with limited agent history information (top row) and multiple future modalities (bottom row).

\vspace{-2mm}
\section{Conclusion}
\label{sec:conclusion}

We introduced~\ourmethod{}, an LLM-style autoregressive behavior model, to better understand the effects of model parameters and dataset size for autonomous driving.
We systematically examined model performance as a function of both dataset size and model capacity, revealing LLM-like scaling laws for data and compute, as well as diminishing returns with increased model size.
We showed the quantitative and qualitative benefits of scaling for planning in real-world driving scenarios.
Additionally, we demonstrated our method on a public motion prediction benchmark, where~\ourmethod{} outperformed state-of-the-art baselines and achieved improved performance through pretraining on a large-scale dataset.

\newpage
\section*{Acknowledgments}
We thank, in alphabetical order,
Abbas Shikari,
Akshay Rangesh,
Allan Lazarovici,
Andy Zhuang,
Ann Yan,
Anton Spiridonov,
Burkay Donderici,
David Fang,
Dennis Park,
Florent Bekerman,
Greg Meyer,
Hans Zhang,
Haytham Mohamed,
Joe Liu,
Jonathan Sorg,
Junyuan Shi,
Ling Li,
Mao Ye,
Matthew Creme,
Peng Tang,
Poonam Suryanarayan,
Rizwan Chaudhry,
Rosalie Zhu,
Siva Karthik Mustikovela,
Srinidhi Myadaboyina,
Tanmay Agarwal,
Xinlei Pan,
Yi Hao,
Yunlong He,
Zaiwei Zhang,
Zisu Dong,
and many others from Cruise data and infra teams,
for their contributions to the paper.

\section*{Impact Statements}
\label{sec:impact}
This paper aims to advance research in behavior modeling for autonomous vehicles.
Since AVs interact with the public, our work has the potential to enhance their behavior around other road users, ultimately contributing to safer roads.

\bibliography{main}
\bibliographystyle{icml2025}

\newpage
\appendix

\clearpage

\appendix
\section{Training Detail}
\label{sec:training_detail}

\subsection{Internal Evaluation: AV Planning}
We train our models on the internal research dataset for 1 epoch, as customary in the LLM scaling literature~\cite{kaplan2020scaling,hoffmann2022training}.

Our models follow the implementations described in~\cite{nayakanti2023wayformer,seff2023motionlm}, and are trained using a batch size of 2048 and a standard Adam optimizer adopted in~\cite{kaplan2020scaling}. We follow the optimal learning rate schedule discovered in~\cite{hoffmann2022training}, which applies a cosine decay with a cycle length equivalent to the total number of training steps. All models were trained on 16 NVIDIA H100 GPUs.

\subsection{External Evaluation: Motion Prediction}
We train our models on the WOMD data following the same setup in~\cite{shi2022motion}. More specifically, we use a batch size of 80 and an AdamW optimizer with a learning rate of 0.0001. The models are trained for 30 epochs, where the learning rate is decayed by a factor of 0.5 every 2 epochs, starting from epoch 20.

\section{Additional Ablation Studies}
\label{sec:scaling_detail}


\subsection{Ablation Study on Attention Heads}
\label{sec:scaling_attention_heads}
\begin{figure}[t]
    \centering
    \includegraphics[width=1\linewidth]{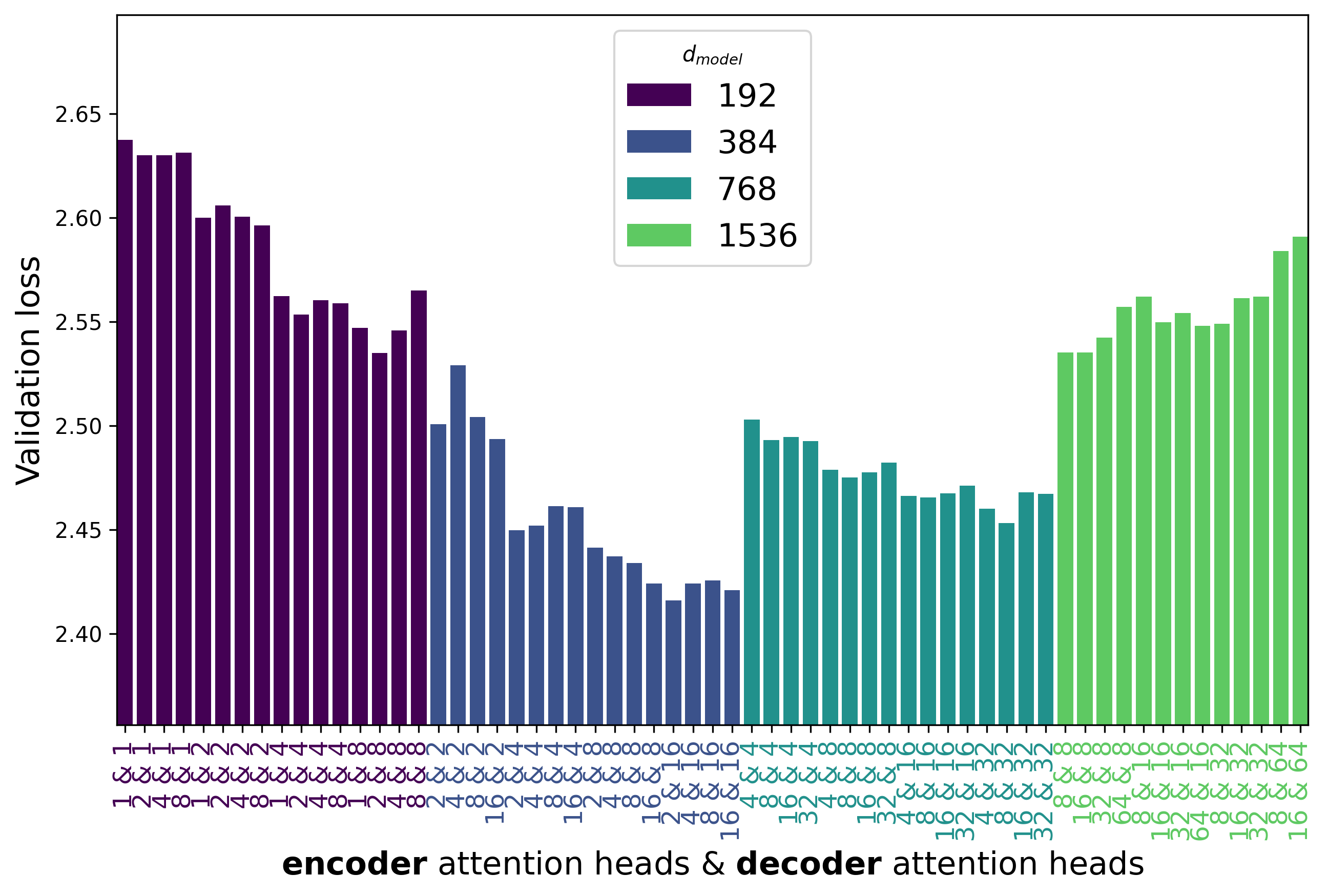}
    \caption{Adjusting the number of attention heads has a smaller effect on the trend of validation losses, compared to changing the hidden dimension.}
    \label{fig:attention_head_scaling}
\end{figure}

In Fig.~\ref{fig:attention_head_scaling}, we present an ablation study examining the impact of the number of heads in the encoder and decoder transformers, while keeping the hidden dimension the same, as specified in Table.~\ref{tab:model_scaling}. 
For each hidden dimension ($d_{\text{model}}$), we conduct experiments across multiple configurations\footnote{For the last group, we omit a few models due to memory constraints.} of attention heads and present their validation losses using a distinct color. While we notice a pattern where more decoder attention heads lead to better performance for small models, the scaling trend is more influenced by the hidden dimension size than by the number of attention heads when the hidden dimension remains fixed. 


\begin{figure}[h!]
    \centering
    \includegraphics[width=1\linewidth]{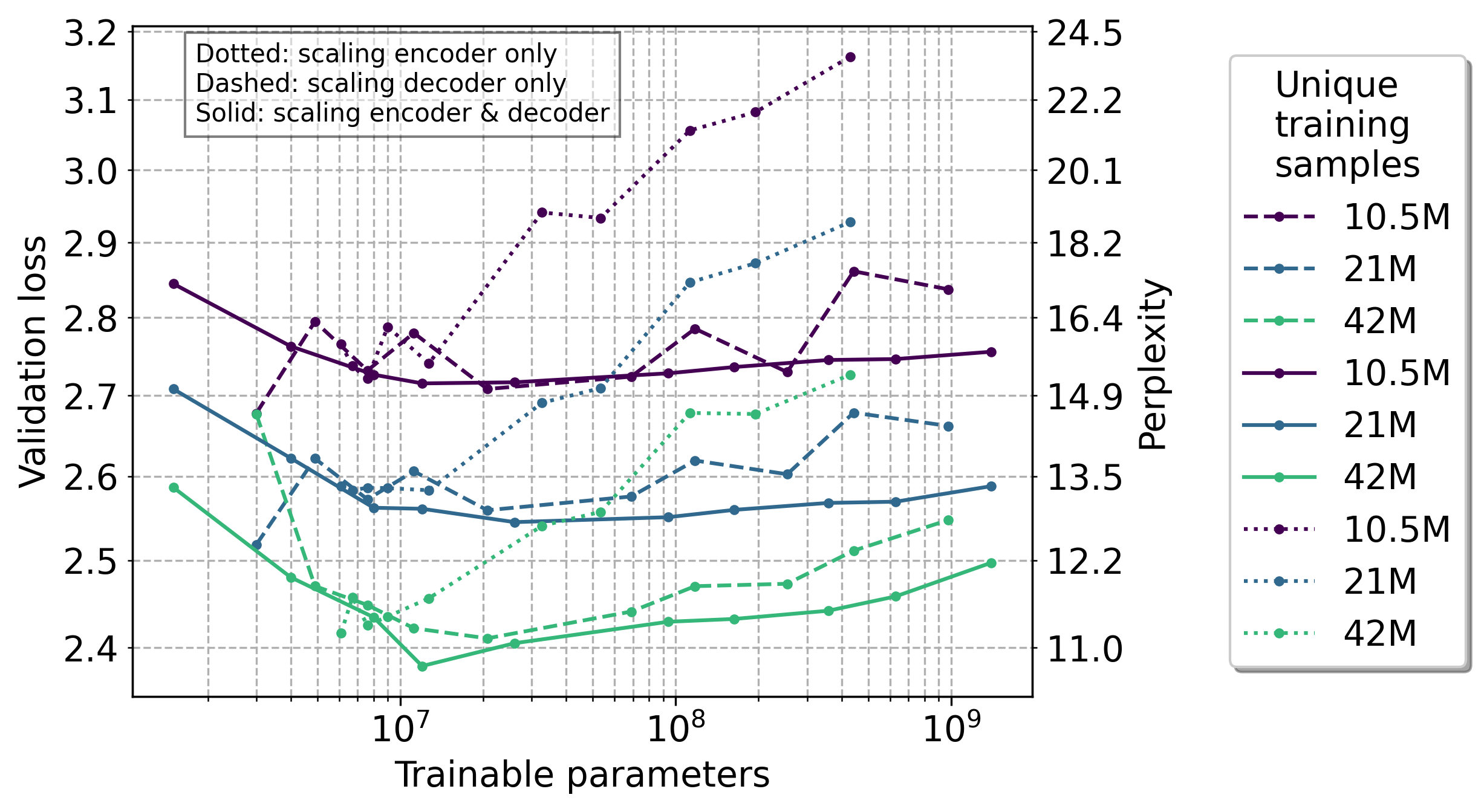}
    \caption{Scaling only the decoder exhibits a similar trend but worse results compared to scaling both encoder and decoder. Scaling only the encoder has little impact beyond 10M parameters.}
    \label{fig:scaling_decoder}
\end{figure}

\subsection{Ablation Study on Decoder Scaling}
In Fig.~\ref{fig:scaling_decoder}, we present model scaling results (in dashed lines) where we only scale up the autoregressive decoder, while keeping the encoder at a fixed size. Compared to scaling both encoder and decoder (in solid lines), scaling only the decoder exhibits a similar trend but yields worse performance, especially for models exceeding 10M parameters in two larger datasets. Therefore, our main results focus on scaling both encoder and decoder.

\subsection{Ablation Study on Encoder Scaling}
In Fig.~\ref{fig:scaling_decoder}, we present model scaling results (in dotted lines) where we only scale up the encoder, while keeping the decoder at a fixed size. The results indicate that encoder scaling is not as effective as decoder scaling.

\section{Scaling From a Perplexity View}
\begin{figure}[t]
    \centering
    \includegraphics[width=1\linewidth]{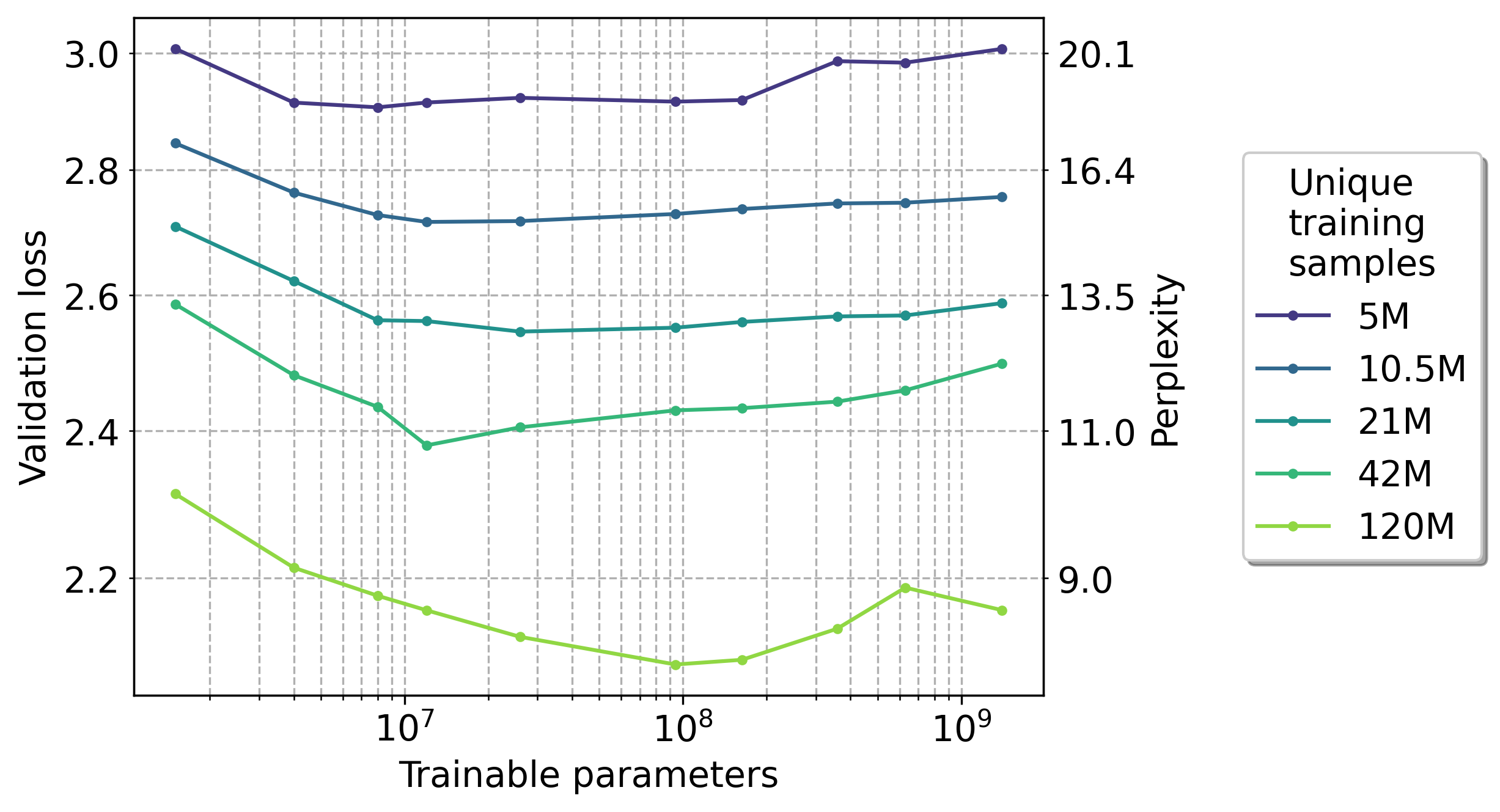}
    \caption{Fig.~\ref{fig:scaling_model} replotted with a perplexity scale on the right $y$-axis for ease of reference.} 
    \label{fig:scaling_model_perplexity}
\end{figure}

\label{sec:scaling_perplexity}
\begin{figure}[t]
    \centering
    \includegraphics[width=0.8\linewidth]{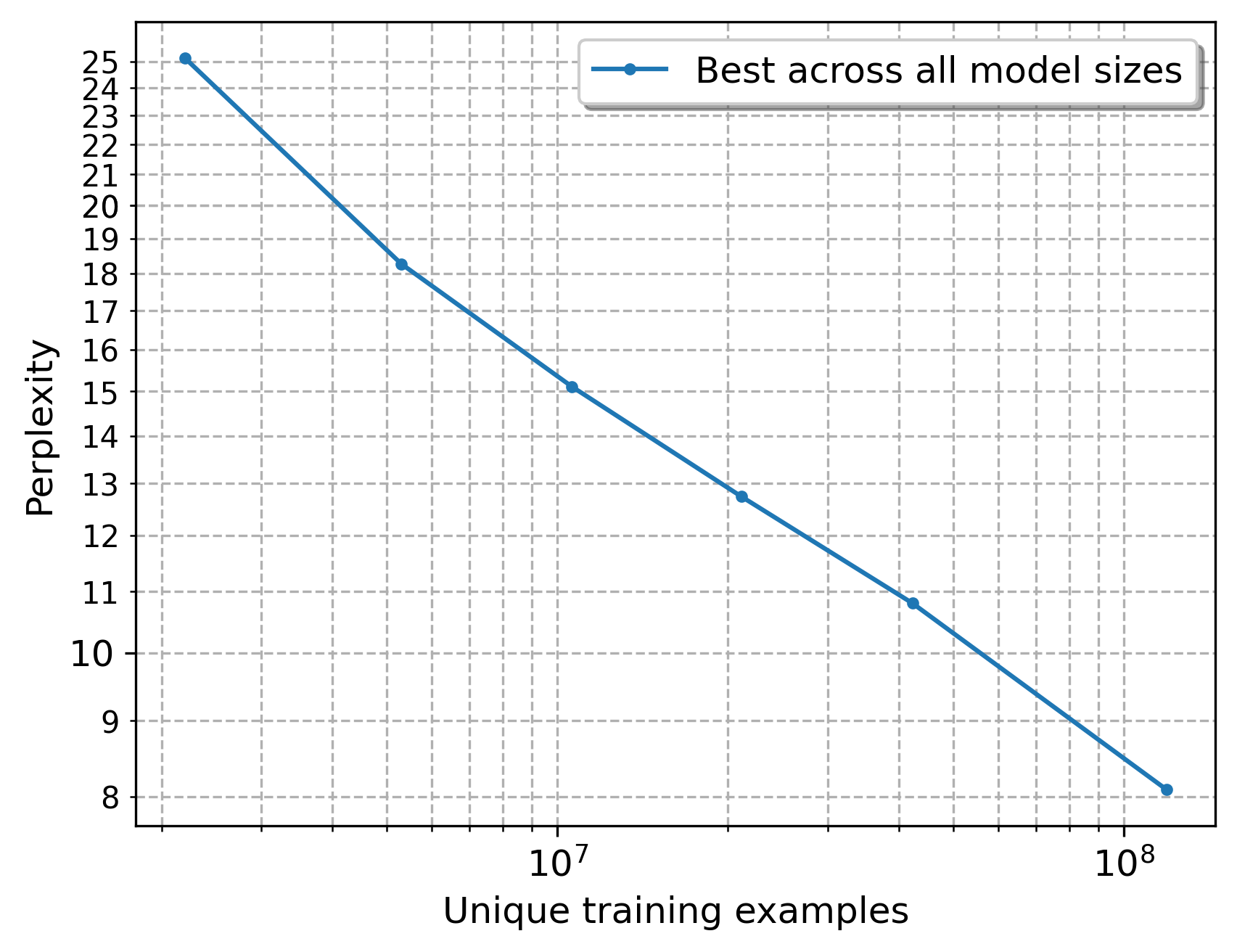}
    \caption{We fit the best perplexity values from all models in Fig.~\ref{fig:scaling_data} as a function of dataset size. A simple linear regression yields $\log(P) = -0.28 \log(D) + 7.22$ with an $R^2$ value of 0.994, where $P$ is the validation perplexity and $D$ is the number of unique training examples.} 
    \label{fig:perplexity_vs_data}
\end{figure}

Following the LLM scaling literature, we present our scaling results as a function of validation loss. 
Another key aspect to highlight is the perplexity scale. Perplexity, defined as the exponentiation of entropy, quantifies how well a probability model predicts a sample. 
It reflects the effective number of choices the model considers: a perplexity of $k$ implies the model is $k$-way perplexed.

In Fig.~\ref{fig:scaling_model_perplexity}, we re-plot the scaling results, adding a secondary $y$-axis to represent perplexity. 
In Fig.~\ref{fig:perplexity_vs_data}, we plot the perplexity of our best models as a function of unique training examples. Extrapolating from the perplexity fit in the figure, a perplexity of 7 would require about $70$M more training examples and a perplexity of 5 would require $500$M more training examples.

\section{Additional Qualitative Examples}
\label{sec:extra_qualitative}

\subsection{Qualitative Comparison of Large Models}
\label{sec:extra_qualitative_large_models}
\begin{figure}[t]
  \centering
    \begin{subfigure}
        \centering
        \includegraphics[width=1\linewidth,trim=0 5pt 0 25pt,clip]{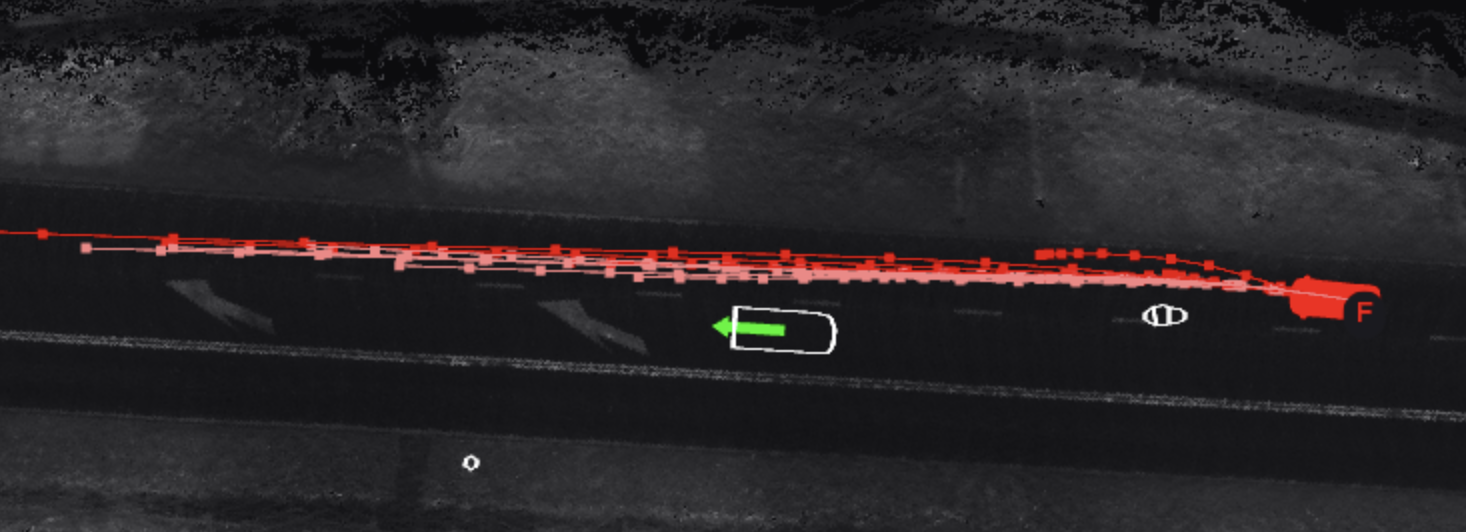}
    \end{subfigure} \vspace{-4mm}\\
    \begin{subfigure}
        \centering
        \includegraphics[width=1\linewidth,trim=0 25pt 0 75pt,clip]{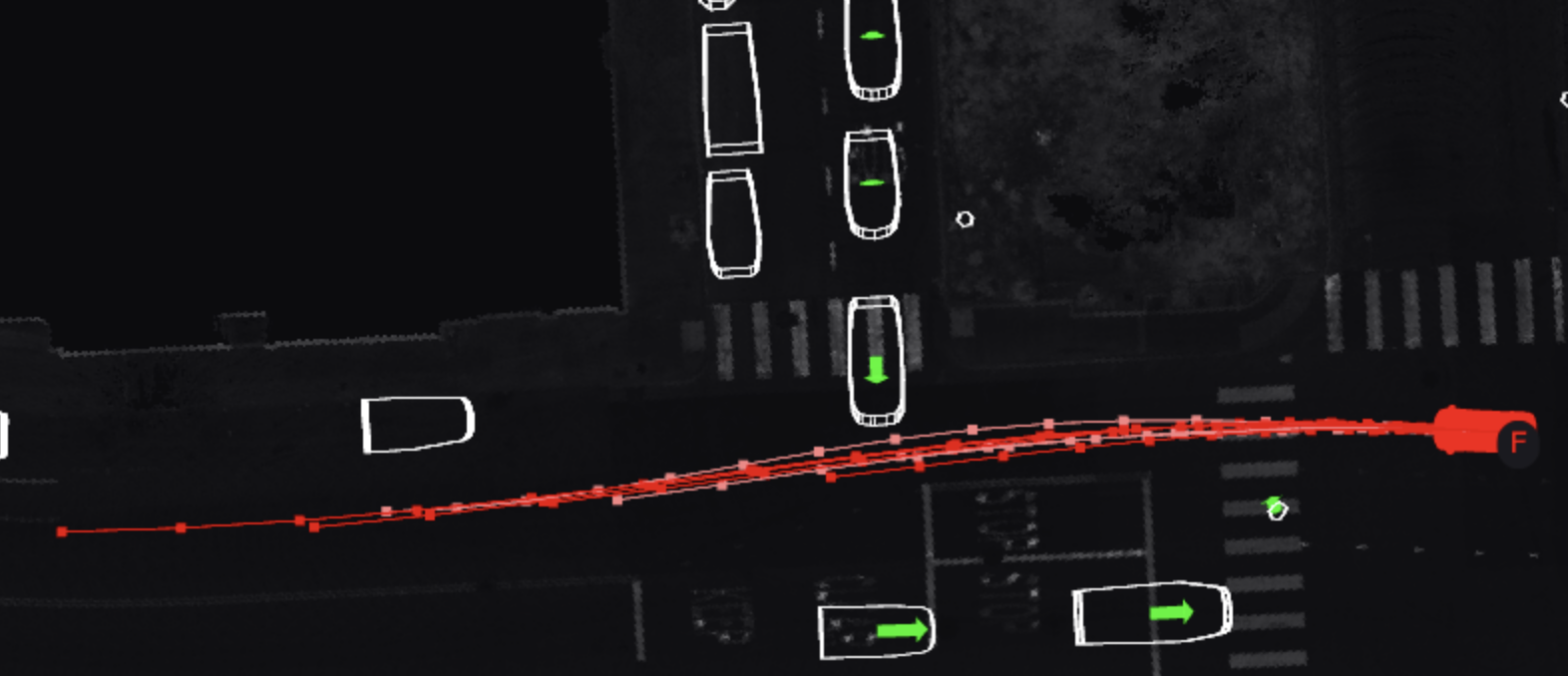}
    \end{subfigure}
    \caption{A larger~\ourmethod{} model at 163M parameters generates trajectories (red) that maintain a safer distance from other traffic agents and road debris compared to a smaller model at 94M parameters (pink).}
    \label{fig:model_scaling_qualitative_large}
\end{figure}

In Fig.~\ref{fig:model_scaling_qualitative_large}, we present two qualitative examples demonstrating that a larger \ourmethod{} model (163M parameters) exhibits better collision avoidance capabilities, compared to its smaller 94M-parameter variant.

\begin{figure}[t]
  \centering
    \includegraphics[width=1\linewidth]{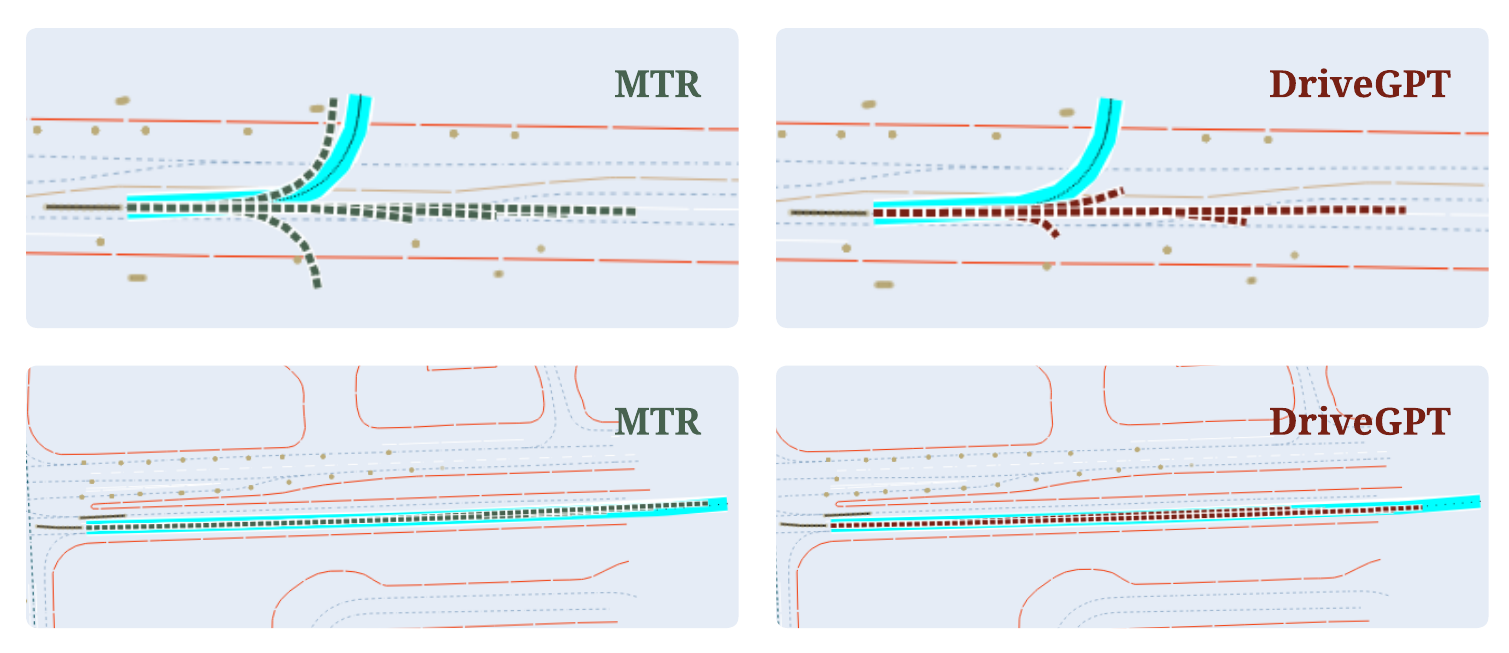}
    \caption{In examples where~\ourmethod{} has the largest error regressions compared to MTR, it predicts less accurate trajectories due to the presence of lane boundary (top row) or missing map information (bottom row). Blue represents ground truth future.}
    \label{fig:womd_worse}
\end{figure}

\begin{table}[t]
  \centering
  \setlength{\tabcolsep}{3pt}
  \begin{tabular}{l|l|cccc}
  \toprule
     Vehicle & Data & mADE$\downarrow$ & mFDE$\downarrow$ & MR$\downarrow$ \\
     \midrule
    \ourmethod{}-WOMD & N/A & 0.6381 & 1.2846 & 0.1233 \\
    \ourmethod{}-Finetune & 40M & 0.6368 & 1.2788 & 0.1217 \\
    \ourmethod{}-Finetune & 80M & 0.6341 & 1.2736 & 0.1207 \\
    \ourmethod{}-Finetune & 120M & \textbf{0.6310} & \textbf{1.2655} & \textbf{0.1195} \\
    \bottomrule
  \end{tabular}
  \caption{Using more training samples during pretraining gives better results (averaged over 3 horizons) in the finetuned model. This verifies the benefit of data scaling as shown in Sec.~\ref{sec:scaling_data}.}
  \label{tab:womd_ablation}
\end{table}

\begin{table}[t]
  \centering
  \setlength{\tabcolsep}{3pt}
  \begin{tabular}{c|cccc}
  \toprule
      Model Parameters & mADE$\downarrow$ & mFDE$\downarrow$ & MR$\downarrow$ \\
     \midrule
    2M & 0.6953 & 1.4297 & 0.1489 \\
    4M & 0.6534 & 1.3202 & 0.1298 \\
    9M & 0.6423 & 1.2955 & 0.1249 \\
    14M & \textbf{0.6381} & \textbf{1.2846} & 0.1233 \\
    21M & 0.6396 & 1.2871 & \textbf{0.1225} \\
    \bottomrule
  \end{tabular}
  \caption{Scaling model parameters in~\ourmethod{}-WOMD improves vehicle prediction metrics up to 14M model parameters. Results are averaged over 3 prediction horizons.}
  \label{tab:womd_model_scaling}
\end{table}

\subsection{Additional Closed-Loop Driving Examples}
In the supplementary video at \url{https://youtu.be/-hLi44PfY8g}, we present additional examples of deploying~\ourmethod{} as a real-time motion planner in a closed-loop setting. The video covers representative challenging scenarios in driving, including a) unprotected left turn, b) double parked vehicle, c) construction zone, d) blow-through cyclist, and e) lane change in heavy traffic. 

\subsection{Failure Cases on WOMD}
We show two representative failure examples of~\ourmethod{} in Fig.~\ref{fig:womd_worse}, selected based on the largest minFDE regressions of our method compared to the MTR baseline on WOMD. In the first example (top row), our method fails to predict trajectories that make a left turn, likely due to the presence of a lane boundary. In the second example (bottom row),~\ourmethod{} predicts an undershooting trajectory that results in a larger longitudinal error due to missing map information in the data.

\section{Additional WOMD Validation Results}
In this section, we present additional ablation studies on the WOMD validation set, due to the submission limit of the WOMD test server.

\subsection{WOMD Data Scaling in Pretraining} 
We validate the effectiveness of data scaling, by pretraining~\ourmethod{} on various sizes of our internal dataset. 
The results in Table~\ref{tab:womd_ablation} indicate that pretraining on more unique samples leads to better results in the finetuned model, aligning with our findings in Sec.~\ref{sec:scaling_data} that data scaling improves model performance.

\subsection{WOMD Model Scaling}
We present an ablation study on model scaling using~\ourmethod{}-WOMD in Table~\ref{tab:womd_model_scaling}. The results show that the model performance continues to improve up to 14M parameters, with no further gains beyond this point due to the limited sample size in WOMD. Therefore, we choose to report the results from the 14M model in Sec.~\ref{sec:external_eval}.


\end{document}